\documentclass{article}
\usepackage{ijcai26}

\usepackage{times}
\usepackage{soul}
\usepackage{url}
\usepackage[hidelinks]{hyperref}
\usepackage[utf8]{inputenc}
\usepackage[small]{caption}
\usepackage{graphicx}
\usepackage{subcaption}  % 子图
\usepackage{amssymb} % 必须包含此包以正常显示 \blacksquare 和 \diamond
\usepackage{amsmath}
\usepackage{amsthm}
\usepackage{booktabs}
\usepackage{algorithm}

\usepackage{algpseudocode} % 提供 \State \For \Return 等

\usepackage[switch]{lineno}
\usepackage{multirow}
\usepackage{makecell}
\usepackage{array}
\usepackage{adjustbox}
\usepackage{tabularx}
\usepackage{float}

\newcolumntype{Y}{>{\centering\arraybackslash}X}

\begin{document}

% Comment out this line in the camera-ready submission
%\linenumbers

\urlstyle{same}

% the following package is optional:
%\usepackage{latexsym}

% See https://www.overleaf.com/learn/latex/theorems_and_proofs
% for a nice explanation of how to define new theorems, but keep
% in mind that the amsthm package is already included in this
% template and that you must *not* alter the styling.
\newtheorem{example}{Example}
\newtheorem{theorem}{Theorem}

% Following comment is from ijcai97-submit.tex:
% The preparation of these files was supported by Schlumberger Palo Alto
% Research, AT\&T Bell Laboratories, and Morgan Kaufmann Publishers.
% Shirley Jowell, of Morgan Kaufmann Publishers, and Peter F.
% Patel-Schneider, of AT\&T Bell Laboratories collaborated on their
% preparation.

% These instructions can be modified and used in other conferences as long
% as credit to the authors and supporting agencies is retained, this notice
% is not changed, and further modification or reuse is not restricted.
% Neither Shirley Jowell nor Peter F. Patel-Schneider can be listed as
% contacts for providing assistance without their prior permission.

% To use for other conferences, change references to files and the
% conference appropriate and use other authors, contacts, publishers, and
% organizations.
% Also change the deadline and address for returning papers and the length and
% page charge instructions.
% Put where the files are available in the appropriate places.

% PDF Info Is REQUIRED.

% Please leave this \pdfinfo block untouched both for the submission and
% Camera Ready Copy. Do not include Title and Author information in the pdfinfo section
\pdfinfo{
/TemplateVersion (IJCAI.2026.0)
}

\title{PMCE: Probabilistic Multi-Granularity Semantics with Caption-Guided Enhancement for Few-Shot Learning}

\author{
Jiaying Wu$^1$
\and
Can Gao$^1$\and
Jinglu Hu$^{2}$\and
Hui Li$^{1}$\and
Xiaofeng Cao$^{3}$\And
Jingcai Guo$^4$\footnote{Corresponding author}\\
\affiliations
$^1$Jiangsu Ocean University, China, 
$^2$Waseda University, Japan\\
$^3$Tongji University, China, 
$^4$The Hong Kong Polytechnic University, Hong Kong SAR\\
\emails
\{jiaying, 2024220911\}@jou.edu.cn, 
jinglu@waseda.jp, 
lih@jou.edu.cn, 
xiaofengcao@tongji.edu.cn, 
jc-jingcai.guo@polyu.edu.hk
}

\maketitle

\begin{abstract}
Few-shot learning aims to identify novel categories from only a handful of labeled samples, where prototypes estimated from scarce data are often biased and generalize poorly. Semantic-based methods alleviate this by introducing coarse class-level information, but they are mostly applied on the support side, leaving query representations unchanged. In this paper, we present PMCE, a \textbf{P}robabilistic few-shot framework that leverages \textbf{M}ulti-granularity semantics with \textbf{C}aption-guided \textbf{E}nhancement. PMCE constructs a nonparametric knowledge bank that stores visual statistics for each category as well as CLIP-encoded class name embeddings of the base classes. At meta-test time, the most relevant base classes are retrieved based on the similarities of class name embeddings for each novel category. These statistics are then aggregated into category-specific prior information and fused with the support set prototypes via a simple MAP update. Simultaneously, a frozen BLIP captioner provides label-free instance-level image descriptions, and a lightweight enhancer trained on base classes optimizes both support prototypes and query features under an inductive protocol with a consistency regularization to stabilize noisy captions. Experiments on four benchmarks show that PMCE consistently improves over strong baselines, achieving up to \textbf{7.71\%} absolute gain over the strongest semantic competitor on MiniImageNet in the 1-shot setting. Our code is available at \url{https://anonymous.4open.science/r/PMCE-275D}.

\end{abstract}

\section{Introduction}
Few-shot learning explores how to recognize novel categories from a small number of labeled samples. Despite the success of deep networks under large-scale supervision, performance often drops sharply in the few-shot regime, where decision boundaries can easily overfit a few supports and fail to transfer to novel classes. Traditional techniques such as Matching Networks~\cite{vinyals2016matching}, Prototypical Networks~\cite{snell2017prototypical}, and MAML~\cite{finn2017model}, together with later strong baselines that revisit pre-training and episodic learning~\cite{chen2019closer,tian2020rethinking,chen2021metabaseline}, have substantially improved the landscape. However, both benchmark studies and recent re-evaluations~\cite{chen2019closer,luo2023closeragain} suggest that the margins over simple baselines are still limited, especially for 1-shot tasks and visually diverse categories.

\begin{figure}[t]
	\centering
	\includegraphics[width=\columnwidth]{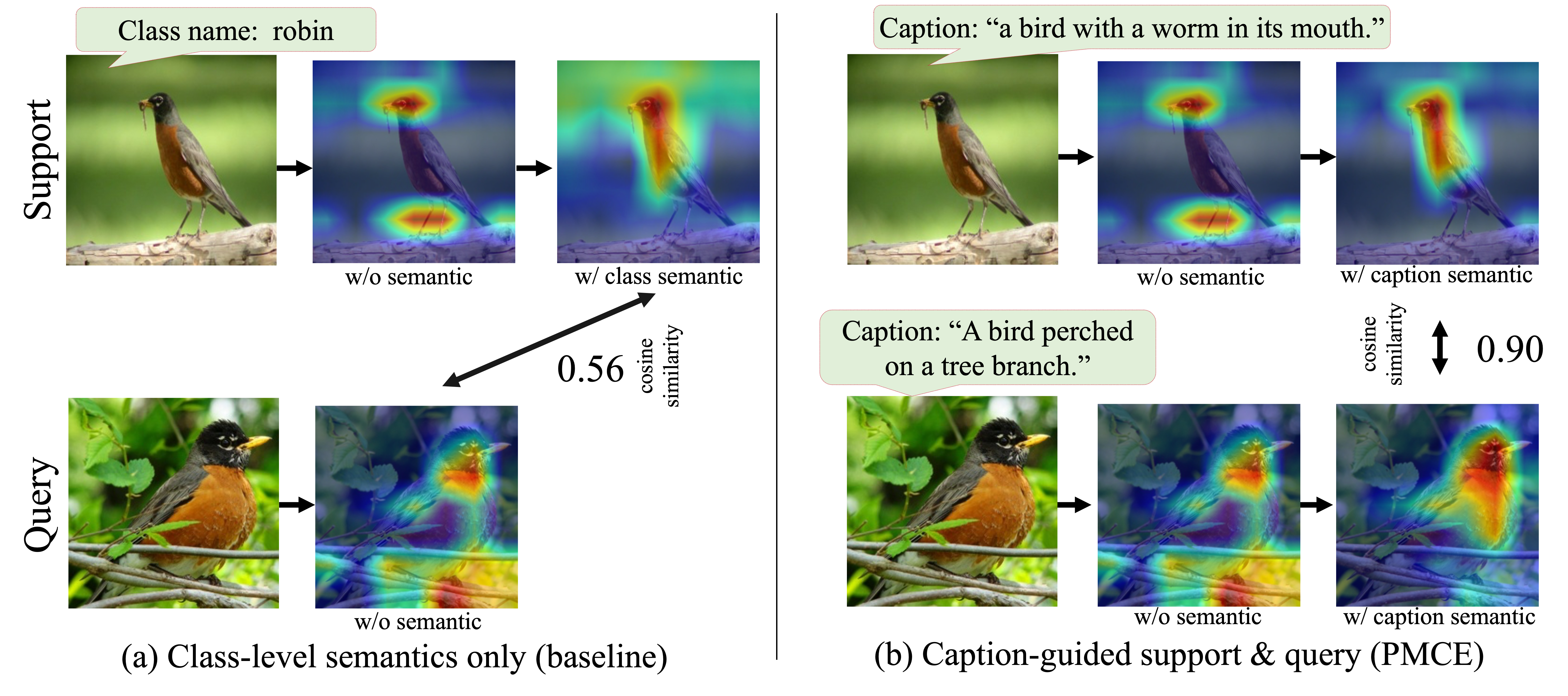}
\caption{
	Comparison of semantic cues for support and query images.
	\textbf{(a) Class-level semantics only (baseline).} The support feature is
	modulated by its class name ``robin'', while the query remains purely visual,
	leading to a relatively weak alignment between them (cosine similarity 0.56).
	\textbf{(b) Caption-guided support and query (PMCE).} BLIP-generated captions
	provide label-free instance-level semantics for both images, yielding more
	focused activation maps and a stronger support–query alignment (cosine
	similarity 0.90).
}

	\label{fig:overview}
\end{figure}

To compensate for the scarcity of labeled samples, semantic-based few-shot learning has become increasingly influential. These methods supplement visual features by introducing auxiliary information such as attributes, class names, human descriptions, and embeddings from visual language models into representations and classifiers~\cite{xing2019adaptive,radford2021learning,DBLP:journals/corr/abs-2303-14123,zhang2024semfew,zhou2025less}. For example, attribute-based and cross-modal approaches adapt visual prototypes with class-level semantics~\cite{xing2019adaptive}. More recently, CLIP-driven prompt-based pipelines encode class names with templates and align them with image features~\cite{radford2021learning,DBLP:journals/corr/abs-2303-14123,zhang2024semfew,DBLP:journals/pr/ChenYTLH26,DBLP:conf/mm/LiW23,dong2025prsn}. It is noteworthy that even lightweight methods using semantics can form strong and robust baselines~\cite{zhou2025less}, which motivates us to consider how semantics should be leveraged to overcome the remaining bottlenecks.

This design choice matters most in the 1-shot scenarios, where prototypes are easily influenced by a single support example and query features often dominate the residual uncertainty. On one hand, distribution-based approaches, such as distribution calibration and MAP-style (Maximum A Posteriori) prototype correction~\cite{yang2021free,wang2023visual,wu2021feature,wu2022improved}, reuse base-class statistics to stabilize novel-class estimation. While effective, their priors are often chosen in the visual space via nearest neighbor search or manually written rules, rather than based on semantic associations between classes. On the other hand, most semantic-based approaches mainly rely on class-level text and inject it into support prototypes or classifier weights~\cite{xing2019adaptive,DBLP:journals/corr/abs-2303-14123,zhang2024semfew}, whereas query features are rarely updated when the model cannot rely on query-set statistics. Therefore, even with strong semantic baselines, performance can be limited by the mismatch between noisy supporting prototypes and purely visual queries.

In this work, we present PMCE, a few-shot framework that uses multi-granularity semantics to address the problems of biased prototypes and weak query representations in 1-shot recognition. PMCE assigns different roles to different semantics. Class-name embeddings are used to identify semantically relevant base categories and turn their statistics into class-specific priors for prototype calibration, while label-free captions provide instance-level cues that refine both support-side prototypes and query features in the same representation space. We train only a lightweight enhancement module on base classes and keep the backbone and vision–language encoders frozen, so improvements mainly come from how semantics are injected rather than from increasing model capacity. Extensive experiments on four benchmarks with two backbones validate the effectiveness of this design, especially in the 1-shot setting. Our contributions are summarized as follows:

\begin{itemize}
	\item We propose a PMCE method that combines semantic-based prior calibration with description-based image representation refinement without requiring adaptation using query set statistics.
	
	\item We leverage class-name semantics to obtain class-specific prior information and enhance the support set and query representation with unlabeled descriptions, thereby constructing a more consistent space for prototype-based inference.
	
	\item We introduce a lightweight consistency regularizer and achieve consistent improvements across four benchmarks using two backbone networks, with the most significant improvement on the 1-shot tasks. 
\end{itemize}

\section{Related Work}
In this section, we present three main streams for existing representative few-shot learning approaches and followed by several methods based on Vision--language models.

\textbf{Metric- and distribution-based few-shot learning.}
Few-shot learning has been extensively explored with techniques like transferable metric spaces and meta-learned initialization. These methods include Matching Networks~\cite{vinyals2016matching}, Prototypical Networks~\cite{snell2017prototypical}, MAML~\cite{finn2017model}, and later strong baselines that revisit pre-training and episodic learning~\cite{chen2019closer,tian2020rethinking,chen2021metabaseline}. Recent work further improves representations and distance functions, such as method FGFL with frequency guidance~\cite{cheng2023frequency} and CPEA with class-aware patch embeddings~\cite{hao2023cpea}. Another line models feature distributions and transfers base-class statistics to novel classes, including distribution calibration~\cite{yang2021distcal} and MAP-style prototype estimation~\cite{wu2021feature,wu2022improved}. 
While effective, they often rely on visual similarities and heuristic selections in the feature space. PMCE follows this calibration perspective, but selects prior candidates with CLIP-encoded class-name semantics and incorporates it with caption-guided feature refinement.

\textbf{Semantic-based few-shot learning.}
Semantic-based methods incorporate semantics like attributes, word vectors, or text embeddings to compensate for limited visual examples~\cite{chen2021semantics,chen2025svip,chen2023zero,chen2022gsmflow,guo2020novel,guo2023graph,guo2024parsnets,guo2021conservative,lu2023decomposed,guo2024multimodal,guo2024fine}. For example, AM3~\cite{xing2019adaptive} adapts visual prototypes with class-level semantics. 
Based on CLIP~\cite{radford2021learning}, BMI~\cite{DBLP:conf/mm/LiW23}, SP-CLIP~\cite{DBLP:journals/corr/abs-2303-14123}, SemFew~\cite{zhang2024semfew} and PRSN~\cite{dong2025prsn} encode class names with prompts and inject text features into metric pipelines or prototype optimization. Most approaches remain class-centric and mainly intervene on support prototypes or classifier weights, while the query representation is often left unchanged or only processed through transduction clues.
In PMCE, class-name semantics guide prior selection, and captions provide instance-level cues to optimize both support and query streams in the same metric space.

\begin{figure*}[t]
	\centering
	\includegraphics[width=0.8\textwidth]{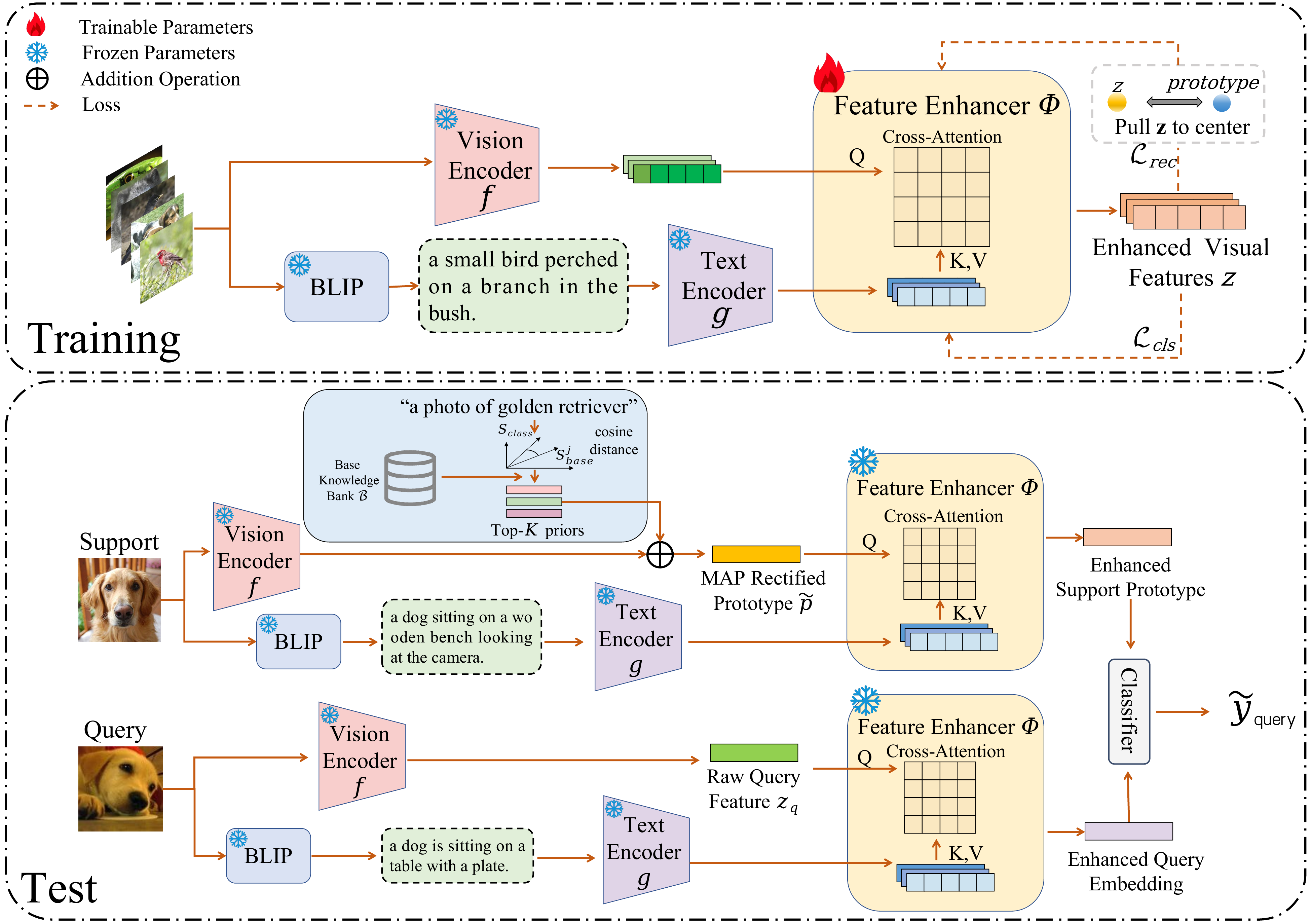}
	\caption{
		Overall framework of PMCE.
		\textbf{Training stage:} On base classes, we build a non-parametric knowledge bank that stores class-wise visual statistics and class-name embeddings, and train a caption-guided enhancer $\Phi$ by fusing visual features from $f$ with BLIP-generated caption semantics from $g$.
		\textbf{Test stage:} (1) extract visual features and label-free captions for both supports and queries;
		(2) use class-name semantics to retrieve related base classes from the knowledge bank and obtain a class-specific prior, which is fused with the support prototype in a MAP-style update to get $\tilde{p}$;
		(3) apply $\Phi$ to enhance $\tilde{p}$ with aggregated support captions and to enhance query features with their own captions, and perform metric-based classification in the aligned feature space.
	}
	\label{fig:framework}
\end{figure*}

\textbf{Vision--language models.}
Vision--language pre-training provides rich textual signals for visual recognition. 
CLIP aligns images and text from large-scale image--caption pairs~\cite{radford2021learning}, and BLIP further improves description generation and integrates visual language representations.~\cite{li2022blip}. Recently, several few-shot work use CLIP textual features to represent category names or manually designed prompts~\cite{DBLP:journals/corr/abs-2303-14123,zhang2024semfew}, while automatic description generation has not been extensively explored under standard inductive evaluation. PMCE leverages BLIP-generated captions as label-free instance semantics and combines them with class-level CLIP semantics, enabling semantics-guided MAP calibration and caption-guided enhancement without using the joint distribution of unlabeled queries.

\section{Preliminaries}

In this section, we explain the few-shot setting and review the MAP-based prototype calibration that our method builds on.

\subsection{Problem Setting}

In the standard few-shot setting, we have a base set $\mathcal{D}_b=\{(x_i,y_i)\}_{i=1}^{|\mathcal{D}_b|}\subset\mathcal{X}\times\mathcal{C}_b$
and a novel set $\mathcal{D}_n=\{(x_i,y_i)\}_{i=1}^{|\mathcal{D}_n|}\subset\mathcal{X}\times\mathcal{C}_n$,
where $\mathcal{C}_b\cap\mathcal{C}_n=\varnothing$. $\mathcal{D}_b$ provides abundant annotated data needed to train the visual backbone and the enhancement module, while $\mathcal{D}_n$ is used only at meta-test time.

At meta-test time, we evaluate the performance on the $N$-way $K$-shot tasks. Each task contains a support set
$\mathcal{S}=\{(x_i,y_i)\}_{i=1}^{N\times K}$ and a query set
$\mathcal{Q}=\{(x_j,y_j)\}_{j=1}^{N\times M}$, where $M$ is the number of query samples per class.
Given support set $\mathcal{S}$ and a query image $x_q$, the goal is to predict its label by
\begin{equation}
\hat{y}_q = \arg\max_{c\in\mathcal{C}} \; p(c\,|\,x_q,\mathcal{S},\mathcal{D}_b),
\label{eq:fsl_objective}
\end{equation}
where $\mathcal{C}$ denotes the set of $N$ classes sampled in the current episode.

\subsection{MAP-based Prototype Calibration}

\label{sec:map_calib}

Following~\cite{wu2021feature,wu2022improved}, we construct a visual knowledge bank based on the priors from base set. Specifically, for each base class $j\in\mathcal{C}_b$, we store
its mean feature
\begin{equation}
\mu_j=\frac{1}{|\mathcal{D}_b^j|}\sum_{(x_i,y_i)\in\mathcal{D}_b^j} f_\theta(x_i),
\end{equation}
where $\mathcal{D}_b^j=\{(x_i,y_i)\in\mathcal{D}_b \mid y_i=j\}$.

For a novel class $c$ in an episode, $p_{\text{init}}^c$ is the empirical prototype calculated from the mean of its support features. Given a class-specific prior mean $\mu_{\text{prior}}^c$ obtained from the knowledge bank, we calibrate the prototype by a MAP estimator under the Gaussian assumptions, which admits a closed-form convex combination:
\begin{equation}
\label{eq:map_pre}
\tilde{p}^c=\alpha\,p_{\text{init}}^c+(1-\alpha)\,\mu_{\text{prior}}^c,
\end{equation}
where $\alpha\in[0,1]$ controls the strength of the prior.

\section{Method}

In this section, we present the semantics-guides prior selection strategy, the caption-guided enhancer, and the training objective in PMCE, the overall framework is shown in Fig.~\ref{fig:framework}.

\subsection{Semantics-Guided Prior Selection}
\label{sec:multi_semantic}

We demonstrate the prior selection with CLIP class-name semantics for MAP prototype
calibration in Eq.~\eqref{eq:map_pre}.

\textbf{Class-level semantics on base classes.}
Let $n_y$ be the natural language name of class $y$. We follow the CLIP-style
prompting strategy and encode
\begin{equation}
	\label{eq:class_semantic}
	s_{\text{class}}^y =
	g_\phi(\text{``a photo of a } n_y\text{''}) \in \mathbb{R}^{d_t},
\end{equation}
which provides a coarse description of category $y$. For each base class $j$, we
store its mean feature $\mu_j$ and class-name embedding
$s_{\text{base}}^j = s_{\text{class}}^j$ in a non-parametric semantic knowledge
bank
\begin{equation}
	\mathcal{B} = \{(\mu_j, s_{\text{base}}^j)\}_{j=1}^{|\mathcal{C}_b|}.
\end{equation}

\textbf{Prior selection for novel classes.}
At meta-test time, for a novel class $c$ with class name $n_c$ and embedding
$s_{\text{class}}^c$, we measure its similarity to each base class via cosine
similarity:
\begin{equation}
	\label{eq:semantic_similarity}
	\text{score}_j =
	\frac{s_{\text{class}}^c \cdot s_{\text{base}}^j}
	{\|s_{\text{class}}^c\|\,\|s_{\text{base}}^j\|}.
\end{equation}
We then select the indices of the top-$K$ base classes to obtain a neighbor set
$\mathcal{N}_{\text{top}}$ and aggregate their visual means into a
class-specific prior. Instead of uniform averaging, we use a similarity-weighted
aggregation:
\begin{equation}
	\label{eq:prior_mean}
	w_k=\frac{\exp(\text{score}_k/\tau)}{\sum_{u\in\mathcal{N}_{\text{top}}}\exp(\text{score}_u/\tau)},
	\qquad
	\mu_{\text{prior}}^c =
	\sum_{k \in \mathcal{N}_{\text{top}}} w_k \mu_k,
\end{equation}
where $\tau$ is a temperature.

\textbf{MAP view with spherical uncertainties.}
We interpret the support prototype $p_{\text{init}}^c$ as a noisy observation of
the true class mean $\mu^c$ under a spherical Gaussian likelihood
$p_{\text{init}}^c \sim \mathcal{N}(\mu^c, \sigma_{\text{like}}^2 \mathbf{I})$,
while the retrieved prior mean provides a spherical Gaussian prior
$p(\mu^c)=\mathcal{N}(\mu_{\text{prior}}^c, \sigma_{\text{prior}}^2 \mathbf{I})$.
Under Gaussian--Gaussian conjugacy, the MAP estimate has a closed form that
reduces to a convex combination:
\begin{equation}
	\label{eq:map_alpha}
	\tilde{p}^c
	=
	\alpha\, p_{\text{init}}^c + (1-\alpha)\, \mu_{\text{prior}}^c,
	\qquad
	\alpha=\frac{\sigma_{\text{prior}}^2}{\sigma_{\text{prior}}^2+\sigma_{\text{like}}^2}.
\end{equation}
Thus, $\alpha$ is determined by the relative confidence between the support
evidence and the retrieved semantic prior. In practice, we use a fixed (or
learnable) scalar $\alpha$ for efficiency, which is consistent with the above
MAP interpretation. Substituting Eq.~\eqref{eq:prior_mean} into
Eq.~\eqref{eq:map_alpha} yields the MAP-calibrated prototype $\tilde{p}^c$ for
class $c$. Compared with \cite{wu2021feature,wu2022improved}, prior selection in
PMCE is driven by CLIP-based class-name semantics rather than by visual distances
or hand-crafted rules.

\subsection{Caption-Guided Dual-Stream Enhancement}
\label{sec:enhancement}

We refine representations with instance-level captions by applying the same
enhancer $\Phi$ to the support and query streams.
Given a visual feature and its caption-derived semantic representation, $\Phi$
outputs an enhanced visual embedding used for prototype construction and query
matching.

\paragraph{Instance-level semantics.}
To provide finer-grained cues and cover unlabeled queries, we use image captions
as instance-level semantics. For any image $x$, we first generate a caption with
a frozen BLIP captioner,
\begin{equation}
	c = \text{BLIP}(x),
\end{equation}
and then encode it with the same text encoder $g_\phi(\cdot)$ to obtain a single
sentence embedding:
\begin{equation}
	s_{\text{inst}} = g_\phi(c)\in\mathbb{R}^{d_t},
\end{equation}
where $d_t$ is the text feature dimension. $s_{\text{inst}}$ is label-free and
thus available for both supports and queries, serving as the semantic input to
$\Phi$ on the two streams.
% and then encode it with the same text encoder $g_\phi(\cdot)$.
% Unlike class-name prompting that yields a single sentence embedding, here we
% retain the token-level hidden states to avoid collapsing the caption into a
% single vector:
% \begin{equation}
% 	\mathbf{S}_{\text{inst}} = g_\phi^{\text{tok}}(c)\in\mathbb{R}^{T\times d_t},
% \end{equation}
% where $T$ is the caption token length (after padding/truncation) and $d_t$ is the
% text feature dimension. $\mathbf{S}_{\text{inst}}$ is label-free and thus
% available for both supports and queries, serving as the semantic input to $\Phi$
% on the two streams.

\paragraph{Enhancement module.}
We define a mapping
\begin{equation}
	\mathbf{v}_{\text{out}} = \Phi(\mathbf{v}_{\text{in}}, \mathbf{S}_{\text{in}}; \Theta),
\end{equation}
where $\mathbf{v}_{\text{in}}\in\mathbb{R}^{1\times d_v}$ is a visual anchor
(e.g., a prototype or a query feature), $\mathbf{S}_{\text{in}}\in\mathbb{R}^{d_t}$
is a semantic context, and $\Theta$ are learnable parameters.
Since visual and textual features live in different spaces, we first project
each semantic token to the visual dimension:
\begin{equation}
    \label{eq:semantic_projection}
	\mathbf{S}_{\text{proj}} =
	\text{ReLU}\!\left(\text{LN}(\mathbf{S}_{\text{in}} W_p +  b_p)\right)
	\in \mathbb{R}^{d_v},
\end{equation}
with $W_p\in\mathbb{R}^{d_t\times d_v}$, $b_p\in\mathbb{R}^{d_v}$ and $\text{LN}(\cdot)$
layer normalization.

We then perform cross-attention by using the visual anchor as the query and the
caption tokens as keys/values:
\begin{equation}
	\mathbf{Q}_{att} = \mathbf{v}_{\text{in}} W_Q,\quad
	\mathbf{K}_{att} = \mathbf{S}_{\text{proj}} W_K,\quad
	\mathbf{V}_{att} = \mathbf{S}_{\text{proj}} W_V,
\end{equation}
where $W_Q,W_K,W_V\in\mathbb{R}^{d_v\times d_k}$ are learnable projections.
The attention output is computed as
\begin{equation}
	\label{eq:attn_token}
\text{Attn}(\mathbf{Q}_{att}, \mathbf{K}_{att}, \mathbf{V}_{att})
= \text{Softmax}\!\left(
\frac{\mathbf{Q}_{att}\mathbf{K}_{att}^\top}{\sqrt{d_k}}
\right) \mathbf{V}_{att}.
\end{equation}
% where $\mathbf{m}\in\mathbb{R}^{1\times T}$ is an additive mask (set to $-\infty$
% on padded positions and $0$ otherwise). This formulation is compatible with
% multi-head attention by applying Eq.~\eqref{eq:attn_token} per head and
% concatenating the outputs.

A residual connection with a learnable scale $\beta$ (initialized to a small
value) preserves the original visual structure:
\begin{equation}
    \label{eq:vout}
	\mathbf{v}_{\text{out}} =
	\mathbf{v}_{\text{in}} + \beta\,
	\text{Attn}(\mathbf{Q}_{att}, \mathbf{K}_{att}, \mathbf{V}_{att}).
\end{equation}

% --- 表格开始 ---
\begin{table*}[t]
	\centering
	
	\caption{Comparison with state-of-the-art few-shot learning methods under the 5-way 1-shot and 5-way 5-shot settings on MiniImageNet and TieredImageNet. All results are reported as mean accuracy (\%) with 95\% confidence intervals. The best performance in each column is highlighted in bold, and ``--'' indicates that the result is not reported.}
	\label{tab:compare_mini_tiered}
	
	% 保持原有的字体设置作为基准
	\fontsize{9pt}{9pt}\selectfont
	\renewcommand{\arraystretch}{1.2}
	\setlength{\tabcolsep}{1pt}
	
	% 使用 resizebox 将表格整体缩小到页面宽度的 90% (0.9\textwidth)
	% 第二个参数 ! 表示高度自动按比例缩放
	\resizebox{0.88\textwidth}{!}{%
		\begin{tabularx}{\linewidth}{%
				>{\raggedright\arraybackslash}m{5.0cm}
				>{\centering\arraybackslash}m{2.0cm}
				>{\centering\arraybackslash}m{1.5cm}
				Y Y Y Y
			}
			\toprule
			\multirow{2}{*}{\centering\textbf{Methods}} &
			\multirow{2}{*}{\centering\textbf{Backbone}} &
			\multirow{2}{*}{\centering\textbf{Semantic}} &
			\multicolumn{2}{c}{\textbf{MiniImageNet}} &
			\multicolumn{2}{c}{\textbf{TieredImageNet}} \\
			\cmidrule(lr){4-5} \cmidrule(lr){6-7}
			& & &
			\textbf{1-shot} & \textbf{5-shot} &
			\textbf{1-shot} & \textbf{5-shot} \\
			\midrule
			
			\multicolumn{7}{l}{\textbf{Metric-based}} \\
			MatchNet~\cite{vinyals2016matching} & ResNet-12 & No & 65.64 ± 0.20 & 78.72 ± 0.15 & 68.50 ± 0.92 & 80.60 ± 0.71 \\
			ProtoNet~\cite{snell2017prototypical} & ResNet-12 & No & 60.37 ± 0.83 & 78.02 ± 0.57 & 53.31 ± 0.89 & 72.69 ± 0.74 \\
			RelationNet~\cite{sung2018learning} & ConvNet64 & No & 52.19 ± 0.83 & 70.20 ± 0.66 & 54.48 ± 0.93 & 71.32 ± 0.78 \\
			SeFeat~\cite{afrasiyabi2022matching} & ResNet-12 & No & 68.32 ± 0.62 & 82.71 ± 0.46 & 73.63 ± 0.88 & 87.59 ± 0.57 \\
			FGFL~\cite{cheng2023frequency} & ResNet-12 & No & 69.14 ± 0.80 & 86.01 ± 0.62 & 73.21 ± 0.88 & 87.21 ± 0.61 \\
			CPEA~\cite{hao2023cpea} & ViT-S & No & 71.97 ± 0.65 & 87.06 ± 0.38 & 76.93 ± 0.70 & 90.12 ± 0.45 \\
			MCNet~\cite{chen2024mutual} & ResNet-18 & No & 72.13 ± 0.43 & 88.20 ± 0.23 & 73.14 ± 0.46 & 87.16 ± 0.32 \\
			Cads~\cite{zhang2024few} & ResNet-12 & No & 66.56 ± 0.19 & 82.74 ± 0.13 & 72.04 ± 0.22 & 86.47 ± 0.15 \\
			SRE-ProtoNet*~\cite{chen2025exploring} & ResNet-12 & No & 70.86 ± 0.44 & 87.01 ± 0.27 & 73.02 ± 0.50 & 87.78 ± 0.31 \\

            % \midrule
			\multicolumn{7}{l}{\textbf{Distribution-based}} \\
			ImprovedMAP~\cite{wu2022improved} & WRN-28-10 & Yes & 71.56 ± 0.74 & 85.05 ± 0.49 & 79.26 ± 0.08 & 90.41 ± 0.72 \\
			TDO~\cite{liu2023capturing} & WRN-28-10 & No & 74.10 ± 0.23 & 86.44 ± 0.13 & 82.68 ± 0.22 & 90.85 ± 0.13 \\
			PBML~\cite{fu2024prototype} & NesT ViT & No & 70.04 ± 0.67 & 85.21 ± 0.58 & 76.73 ± 0.88 & 90.11 ± 0.65 \\
			DDC~\cite{chen2025ddc} & WRN-28-10 & No & 69.17 ± 1.25 & 85.23 ± 0.97 & -- & -- \\
			
			\midrule
			\multicolumn{7}{l}{\textbf{Semantic-based}} \\
			KTN~\cite{peng2019few} & ConvNet & Yes & 64.42 ± 0.72 & 74.16 ± 0.56 & 63.43 ± 0.21 & 74.32 ± 0.58 \\
			AM3~\cite{xing2019adaptive} & ResNet-12 & Yes & 65.30 ± 0.49 & 78.10 ± 0.36 & 69.08 ± 0.47 & 82.58 ± 0.31 \\
            SEGA~\cite{yang2022sega} & ResNet-12 & Yes & 69.04 ± 0.26    & 79.03 ± 0.18 & 72.18 ± 0.30 & 84.28 ± 0.21 \\
			BMI~\cite{DBLP:conf/mm/LiW23} & ResNet-12 & Yes & 77.01 ± 0.34 & 84.85 ± 0.27 & 78.37 ± 0.44 & 86.30 ± 0.32 \\
			MAVSI~\cite{zhao2024multi} & ResNet-12 & Yes & 69.74 ± 0.21 & 82.23 ± 0.41 & 74.61 ± 0.18 & 87.45 ± 0.46 \\
            SEVPro~\cite{DBLP:conf/ijcai/CaiLH024} & ResNet-12 & Yes & 71.81 ± 0.22  & 78.88 ± 0.18  & 72.77 ± 0.30  & 84.04 ± 0.21 \\
			PRSN~\cite{dong2025prsn} & ResNet-12 & Yes & 71.09 ± 0.90 & 86.82 ± 0.61 & 74.14 ± 0.86 & 88.54 ± 0.61 \\

            % \midrule
			\multicolumn{7}{l}{\textbf{Vision–language based}} \\
			SP-CLIP~\cite{DBLP:journals/corr/abs-2303-14123} & ViT-T & Yes & 72.31 ± 0.40 & 83.42 ± 0.30 & 78.03 ± 0.46 & 88.55 ± 0.32 \\
			SP-SBERT~\cite{DBLP:journals/corr/abs-2303-14123} & ViT-T & Yes & 70.70 ± 0.42  & 83.55 ± 0.30 & 73.31 ± 0.50  & 88.56 ± 0.32 \\
            SP-Glove~\cite{DBLP:journals/corr/abs-2303-14123} & ViT-T & Yes & 70.81 ± 0.42  & 83.31 ± 0.30  & 74.68 ± 0.50   & 88.64 ± 0.31 \\
			SemFew-Trans~\cite{zhang2024semfew} & Swin-T & Yes & 78.94 ± 0.66 & 86.49 ± 0.50 & 82.37 ± 0.77 & 89.89 ± 0.52 \\
			SYNTRANS~\cite{DBLP:conf/ijcai/TangHQ25} & ResNet-12 & Yes & 76.20 ± 0.69  & 86.12 ± 0.54  & 79.69 ± 0.81  & 87.78 ± 0.60 \\

			\midrule
			\multirow{2}{*}{\centering\textbf{PMCE(\textbf{Ours})}} &
			ResNet-12 & Yes & 82.90 ± 0.78 & 92.29 ± 0.46 & \textbf{83.50 ± 0.32} & \textbf{95.03 ± 0.34} \\
			& Swin-T & Yes & \textbf{85.03 ± 0.67} & \textbf{92.77 ± 0.37} & 83.13 ± 0.33 & 93.23 ± 0.43 \\

			\bottomrule
		\end{tabularx}%
	}
\end{table*}
% --- 表格结束 ---

% --- 表格开始 ---
\begin{table*}
	\centering
	
	\caption{Comparison with state-of-the-art few-shot learning methods under the 5-way 1-shot and 5-way 5-shot settings on CIFAR-FS and FC100. All results are reported as mean accuracy (\%) with 95\% confidence intervals. The best performance in each column is highlighted in bold, and ``--'' indicates that the result is not reported.}
	\label{tab:compare_cifar_fc100}
	
	\fontsize{9pt}{9pt}\selectfont
	\renewcommand{\arraystretch}{1.2}
	\setlength{\tabcolsep}{1pt}
	
	% 使用 resizebox 将表格整体缩小到页面宽度的 90% (0.9\textwidth)
	% 保持与上一张表格一致的缩放比例
	\resizebox{0.88\textwidth}{!}{%
		\begin{tabularx}{\linewidth}{%
				>{\raggedright\arraybackslash}m{5.0cm}
				>{\centering\arraybackslash}m{2.0cm}
				>{\centering\arraybackslash}m{1.5cm}
				Y Y Y Y
			}
			\toprule
			\multirow{2}{*}{\centering\textbf{Methods}} &
			\multirow{2}{*}{\centering\textbf{Backbone}} &
			\multirow{2}{*}{\centering\textbf{Semantic}} &
			\multicolumn{2}{c}{\textbf{CIFAR-FS}} &
			\multicolumn{2}{c}{\textbf{FC100}} \\
			\cmidrule(lr){4-5} \cmidrule(lr){6-7}
			& & &
			\textbf{1-shot} & \textbf{5-shot} &
			\textbf{1-shot} & \textbf{5-shot} \\
			\midrule
			
			\multicolumn{7}{l}{\textbf{Metric-based}} \\
			ProtoNet~\cite{snell2017prototypical} & ResNet-12 & No & 55.50 ± 0.70 & 72.00 ± 0.60 & 41.54 ± 0.76 & 57.08 ± 0.76 \\
			MetaOptNet~\cite{lee2019meta} & ResNet-12 & No & 72.80 ± 0.70 & 84.30 ± 0.50 & 47.20 ± 0.60 & 55.50 ± 0.60 \\
			CORL~\cite{He_2023_WACV} & ResNet-12 & No & 74.13 ± 0.71 & 87.54 ± 0.51 & 44.82 ± 0.73 & 61.31 ± 0.54 \\
			QSFormer~\cite{wang2023few} & ResNet-12 & No & 46.51 ± 0.26 & 61.58 ± 0.25 & -- & -- \\
			SSFormers~\cite{chen2023sparse} & ResNet-12 & No & 74.50 ± 0.21 & 86.61 ± 0.23 & 43.72 ± 0.21 & 58.92 ± 0.18 \\
			Cads~\cite{zhang2024few} & ResNet-12 & No & 73.23 ± 0.21 & 87.67 ± 0.14 & -- & -- \\

            % \midrule
			\multicolumn{7}{l}{\textbf{Distribution-based}} \\
			PBML~\cite{fu2024prototype} & ResNet-12 & No & 73.07 ± 0.59 & 85.51 ± 0.41 & 47.92 ± 0.49 & 62.96 ± 0.51 \\
			DDC~\cite{chen2025ddc} & WRN-28-10 & No & 76.47 ± 1.86 & 87.24 ± 0.89 & -- & -- \\
			
			\midrule
			\multicolumn{7}{l}{\textbf{Semantic-based}} \\
			
            ProtoNet + SEVPro~\cite{cai2024little} & ResNet-12 & Yes & 74.01 ± 0.29 & 85.86 ± 0.20 & -- & -- \\
			PRSN~\cite{dong2025prsn} & ResNet-12 & Yes & 78.72 ± 0.96 & 90.22 ± 0.60 & 47.91 ± 0.57 & 62.95 ± 0.52 \\

            % \midrule
			\multicolumn{7}{l}{\textbf{Vision–language based}} \\
            LPE~\cite{Yang_2023_WACV} & ResNet-12 & Yes & 74.88 ± 0.45 & 85.30 ± 0.35 & -- & -- \\
			SP-Pretrain~\cite{DBLP:journals/corr/abs-2303-14123} & ViT-S & Yes & 71.99 ± 0.47 & 85.98 ± 0.34 & 43.77 ± 0.39 & 59.48 ± 0.39 \\
			SAFF~\cite{DBLP:conf/cvpr/Cumplido0R25} & Vit-B & Yes & 78.50 ± 0.30 & \textbf{90.26 ± 0.08}  & 47.17 ± 0.12 &66.22 ± 0.51 \\

			\midrule
			\multirow{2}{*}{\centering\textbf{PMCE(\textbf{Ours})}} &
			ResNet-12 & Yes & 76.97 $\pm$ 0.81 & 84.76 ± 0.61 & 49.81 ± 0.82 & 63.67 ± 0.73 \\
			& Swin-T & Yes & \textbf{80.92 $\pm$ 0.72} & 89.02 ± 0.58 & \textbf{52.46 $\pm$ 0.82} & \textbf{67.00 $\pm$ 0.75} \\
			
			\bottomrule
		\end{tabularx}%
	}

\end{table*}

\paragraph{Support and query streams.}
We apply $\Phi(\cdot)$ symmetrically to supports and queries.
For class $c$, let $\mathcal{S}_c$ be its support set and
$\mathbf{S}_{\text{inst}}^{(i)}$ be the caption embedding of support $x_i\in\mathcal{S}_c$. We average support caption embeddings to form a class-level semantic context:
\begin{equation}
	s_{\text{proto}}^c =
	\frac{1}{|\mathcal{S}_c|}
	\sum_{x_i\in\mathcal{S}_c} s_{\text{inst}}^{(i)}
	\in\mathbb{R}^{d_t}.
\end{equation}
We then take the MAP-calibrated prototype $\tilde{p}^c$ as the visual anchor to
obtain the final enhanced prototype:
\begin{equation}
	\mathbf{P}_{\text{final}}^c = \Phi(\tilde{p}^c, \mathbf{S}_{\text{proto}}^c).
\end{equation}
For a query image $x_q$ with visual feature $\mathbf{z}_q = f_\theta(x_q)$ and
its caption embedding $\mathbf{S}_{\text{inst}}^{q}$, we compute
\begin{equation}
	\mathbf{Z}_{\text{final}}^q = \Phi(\mathbf{z}_q, \mathbf{S}_{\text{inst}}^{q}).
\end{equation}
Each query is processed independently, so the overall framework remains strictly
inductive.

\subsection{Training Objective with Consistency Regularization}
\label{sec:training}

We freeze $f_\theta$ and $g_\phi$ and train only the enhancer $\Phi$ together
with an auxiliary classifier on base classes. Let $v_{\text{out}}^{(i)}$ be the
enhanced feature of a training sample $x_i$ with label $y_i$.

\paragraph{Classification and prototype-preserving regularization.}
We use a standard cross-entropy loss
\begin{equation}
	\mathcal{L}_{\text{cls}} =
	-\frac{1}{B} \sum_{i=1}^B
	\log P(y_i \mid v_{\text{out}}^{(i)}),
\end{equation}
and a lightweight prototype-preserving term that discourages excessive drift
from the frozen backbone structure:
\begin{equation}
	\mathcal{L}_{\text{rec}} =
	\frac{1}{B} \sum_{i=1}^B
	\big\| v_{\text{out}}^{(i)} - \mu_{\text{proto}}^{y_i} \big\|_1.
\end{equation}
We use a small $\lambda_{\text{rec}}$ so that the enhancer can still introduce
caption-guided adaptation while keeping base-class geometry stable.

\paragraph{Caption consistency with supervised contrastive learning.}
Caption quality may vary, so we add a consistency regularization on caption
semantics in the projected space. For sample $x_i$, let
$\mathbf{s}_{\text{proj}}^{(i)}\in\mathbb{R}^{d_v}$ be the projected caption
embedding (Eq.~\eqref{eq:semantic_projection}). We normalize it as
\begin{equation}
	v_i^{\text{cap}} = \frac{\mathbf{s}_{\text{proj}}^{(i)}}{\|\mathbf{s}_{\text{proj}}^{(i)}\|}
	\in\mathbb{R}^{d_v}.
\end{equation}
% Caption quality may vary, so we further introduce a consistency regularization
% on caption semantics in the projected space. For sample $x_i$, let
% $\mathbf{S}_{\text{proj}}^{(i)}\in\mathbb{R}^{T\times d_v}$ be the projected
% caption tokens (Eq.~\eqref{eq:semnatic_projection}). We obtain a pooled caption
% representation by token averaging (ignoring padding):
% \begin{equation}
% 	v_i^{\text{cap}} = \text{Pool}\!\left(\mathbf{S}_{\text{proj}}^{(i)}\right)
% 	\in\mathbb{R}^{d_v}.
% \end{equation}
We encourage captions from the same class to be coherent while maintaining
inter-class separation via a supervised contrastive loss:
\begin{equation}
	\mathcal{L}_{\text{con}} =
	\frac{1}{B}\sum_{i=1}^{B}
	\left[
	-\frac{1}{|\mathcal{P}(i)|}\sum_{p\in\mathcal{P}(i)}
	\log
	\frac{
	\exp(\cos(v_i^{\text{cap}}, v_p^{\text{cap}})/\tau_c)
	}{
	\sum\limits_{\substack{a=1\\a\neq i}}^{B}
	\exp(\cos(v_i^{\text{cap}}, v_a^{\text{cap}})/\tau_c)
	}
	\right],
\end{equation}
where $\mathcal{P}(i)=\{p\neq i\mid y_p=y_i\}$ denotes positives in the mini-batch
and $\tau_c$ is a temperature.

\paragraph{Overall objective.}
The final objective is
\begin{equation}
	\mathcal{L}_{\text{total}} =
	\mathcal{L}_{\text{cls}}
	+ \lambda_{\text{rec}} \mathcal{L}_{\text{rec}}
	+ \lambda_{\text{con}} \mathcal{L}_{\text{con}},
\end{equation}
with $\lambda_{\text{rec}}$ and $\lambda_{\text{con}}$ tuned on the validation
set. During meta-test, $f_\theta$, $g_\phi$ and $\Phi$ are all frozen, and classification is performed with logistic regression using enhanced support and
query features.

\section{Experiments}

\textbf{Datasets.}
We evaluate PMCE on four standard few-shot benchmarks, including MiniImageNet, TieredImageNet, CIFAR-FS, and FC100.
MiniImageNet~\cite{vinyals2016matching} and TieredImageNet~\cite{ren2018meta} are subsets of ImageNet~\cite{russakovsky2015imagenet} with 64, 16, and 20 classes for base, validation, and novel splits, and 351, 97, and 160 classes for base, validation, and novel splits, respectively.
CIFAR-FS~\cite{bertinetto2018meta} follows a random split of CIFAR-100 into 64, 16, and 20 classes, while FC100~\cite{oreshkin2018tadam} adopts a superclass-based split with 60, 20, and 20 classes to reduce semantic overlap between base and novel categories.

\textbf{Implementation details.}
Following~\cite{zhao2024multi}, we use ResNet-12~\cite{he2016deep} and Swin-T~\cite{liu2021swin} as visual backbones.
Captions are generated using a frozen BLIP captioner, and both class names and captions are encoded by ViT-B/16 CLIP~\cite{radford2021learning} into 512-dimensional semantic features. All CLIP and BLIP components remain frozen throughout training and evaluation. The enhancer $\Phi$ consists of a projection layer with Linear, LayerNorm, and ReLU, followed by multi-head attention, with a residual update scaled by a learnable factor $\beta$ initialized to 0.1. We use $h=4$ heads and dropout 0; the key dimension is $d_k=160$ for ResNet-12 ($d_v=640$) and $d_k=192$ for Swin-T ($d_v=768$).
All backbones and BLIP/CLIP modules remain frozen, and inference is strictly inductive.

% We tune the temperatures $\tau$ (Eq.~\eqref{eq:prior_mean}) and $\tau_c$ on the validation split and observe that the method is not sensitive within a reasonable range.

We follow a two-stage training protocol. We load the backbone pre-trained in~\cite{zhao2024multi}, freeze it, and train only $\Phi$ together with an auxiliary classifier for 50 epochs using Adam with learning rate $10^{-4}$ and batch size 128. During evaluation, we sample 600 episodes for each 5-way 1-shot and 5-way 5-shot setting and report mean accuracy with 95\% confidence intervals; logistic regression is used by default, and nearest-prototype yields similar trends. Unless otherwise specified, we set $\alpha=0.33$ for 1-shot and $\alpha=0.7$ for 5-shot in Eq.~(\ref{eq:map_pre}), use top-$k$ retrieval with $k=7$ and temperature $\tau=1.0$, and set $\tau_c=0.1$, $\lambda_{\text{rec}}=1.0$, and $\lambda_{\text{con}}=1.0$ for training.We use class-name embeddings for retrieval since they are stable and class-consistent, while captions can be noisier on low-resolution images.
We also experimented with token-level caption features, but they lead to significantly larger intermediate files and storage overhead, so we report sentence-embedding results by default.

% Captions are generated offline using a frozen BLIP captioner and are reused for all episodes.
% We cache both captions and their CLIP text embeddings, so meta-test does not access query-set statistics beyond per-sample caption encoding.

% All CLIP and BLIP components remain frozen throughout training and evaluation.
% The attention module in $\Phi$ uses four heads, and the output dimension is matched to the backbone feature dimension. We use cosine similarity for prototype-based classification, consistent with the retrieval metric used in Eq.~(\ref{eq:semantic_similarity}). All hyperparameters are selected on the validation split and then fixed for novel-class evaluation. We repeat experiments with fixed random seeds and report the average over sampled episodes.

\begin{table}[t]
	\centering
	\caption{Ablation of PMCE components under the 5-way 1-shot setting
		on MiniImageNet and FC100 with the Swin-T backbone.
		MAP denotes semantics-guided prior calibration; 
		Enh.(S) and Enh.(Q) denote caption-guided enhancement on the support
		and query streams. We report mean accuracy (\%) $\pm$ 95\% confidence intervals.}
	\label{tab:ablation_main}
	\small
	\begin{tabular}{ccc cc}
		\toprule
		MAP & Enh.(S) & Enh.(Q) 
		& MiniImageNet
		& FC100 \\
		\midrule
		$\times$ & $\times$ & $\times$ 
		& $71.02 \pm 0.79$ 
		& $45.27 \pm 0.75$ \\
		$\times$ & $\checkmark$ & $\times$ 
		& $72.18 \pm 0.78$ 
		& $46.03 \pm 0.74$ \\
		$\times$ & $\times$ & $\checkmark$ 
		& $72.49 \pm 0.81$ 
		& $45.47 \pm 0.77$ \\
		$\checkmark$ & $\times$ & $\times$ 
		& $74.93 \pm 0.76$ 
		& $45.36 \pm 0.75$ \\
		$\checkmark$ & $\checkmark$ & $\times$ 
		& $74.99 \pm 0.74$ 
		& $45.15 \pm 0.77$ \\
		$\checkmark$ & $\times$ & $\checkmark$ 
		& $77.17 \pm 0.76$ 
		& $45.27 \pm 0.78$ \\
		$\times$ & $\checkmark$ & $\checkmark$ 
		& $79.42 \pm 0.73$ 
		& $48.79 \pm 0.80$ \\
		\midrule
		$\checkmark$ & $\checkmark$ & $\checkmark$ 
		& $\mathbf{85.03 \pm 0.67}$ 
		& $\mathbf{52.46 \pm 0.82}$ \\
		\bottomrule
	\end{tabular}
\label{tab:ablation_7rows}
\end{table}

\begin{figure}[ht]
	\centering
	\begin{subfigure}{0.23\textwidth}
		\centering
		\includegraphics[width=\textwidth]{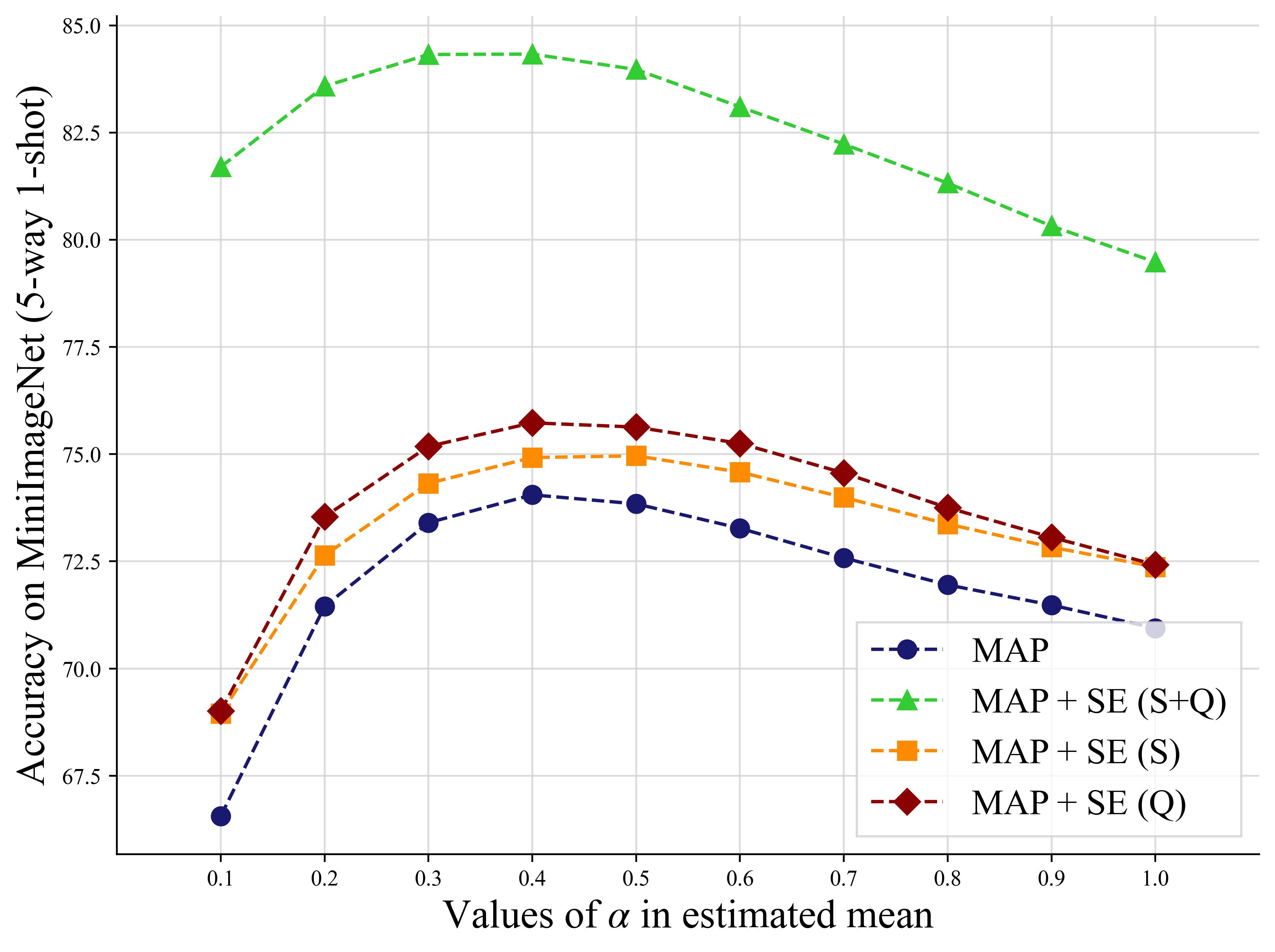}
		\caption{}
		\label{fig:alpha_1shot}
	\end{subfigure}
	\hfill
	\begin{subfigure}{0.23\textwidth}
		\centering
		\includegraphics[width=\textwidth]{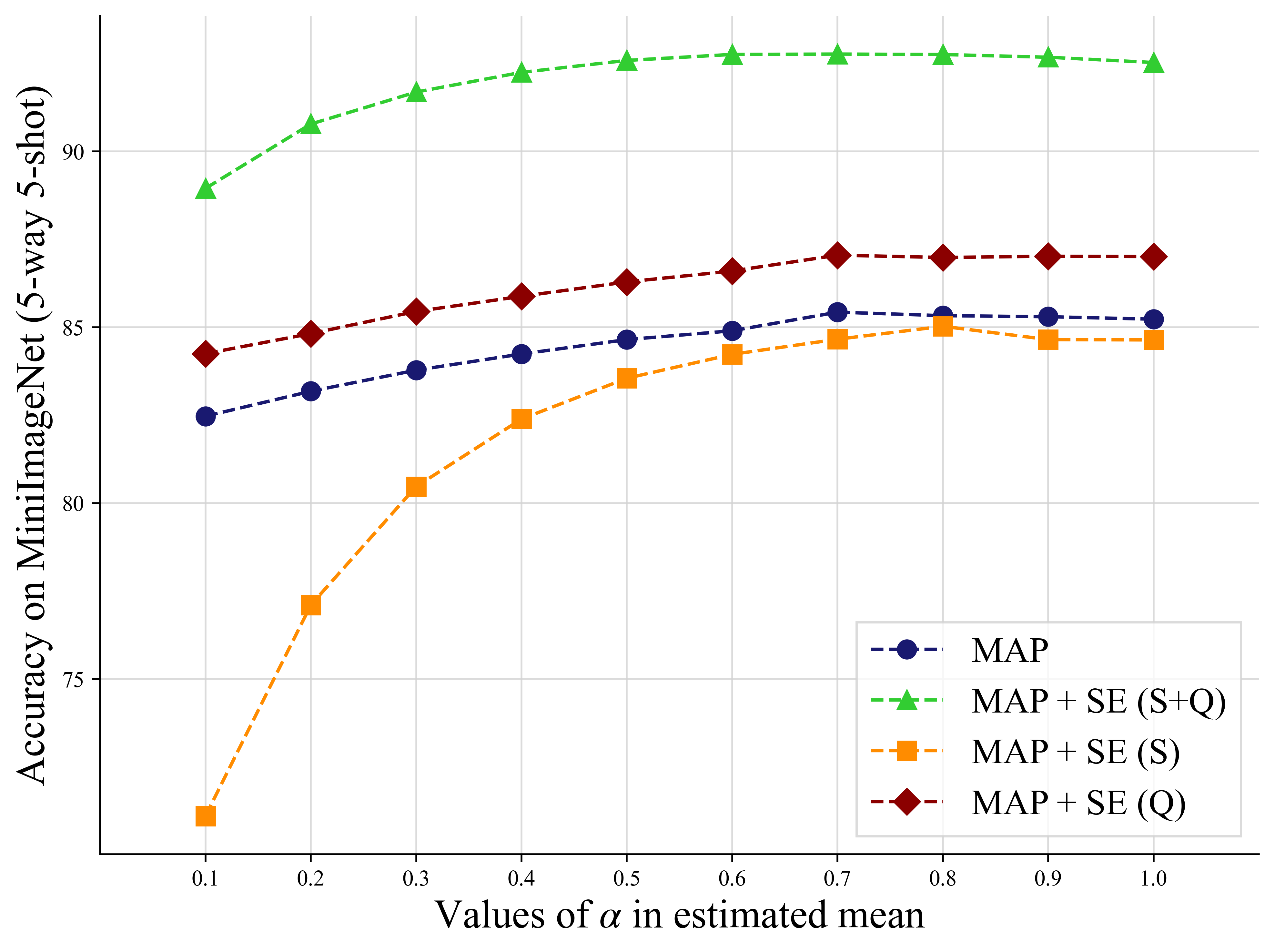}
		\caption{}
		\label{fig:alpha_5shot}
	\end{subfigure}
	\hfill
	\begin{subfigure}{0.23\textwidth}
		\centering
		\includegraphics[width=\textwidth]{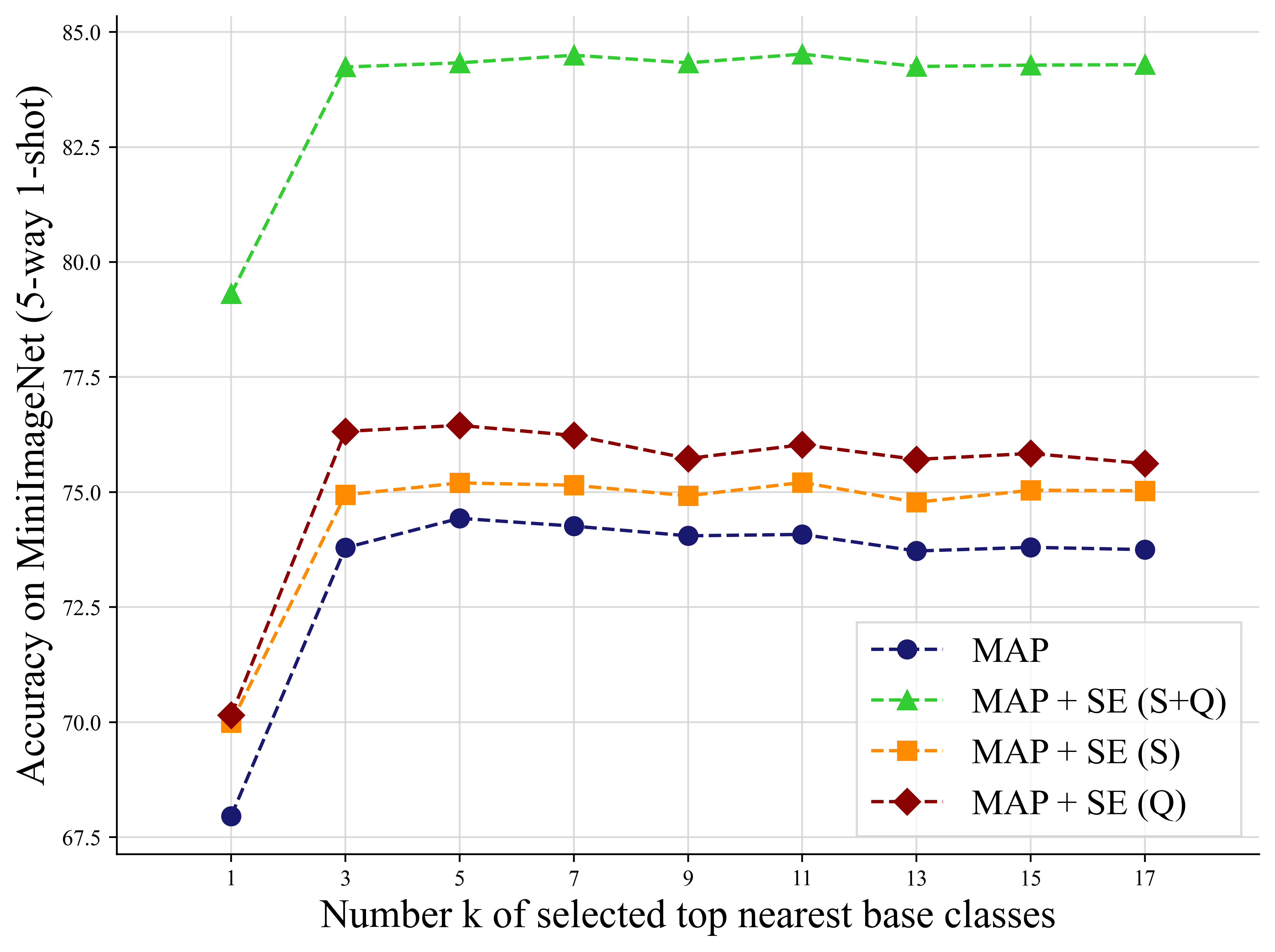}
		\caption{}
		\label{fig:topk_1shot}
	\end{subfigure}
	\hfill
	\begin{subfigure}{0.23\textwidth}
		\centering
		\includegraphics[width=\textwidth]{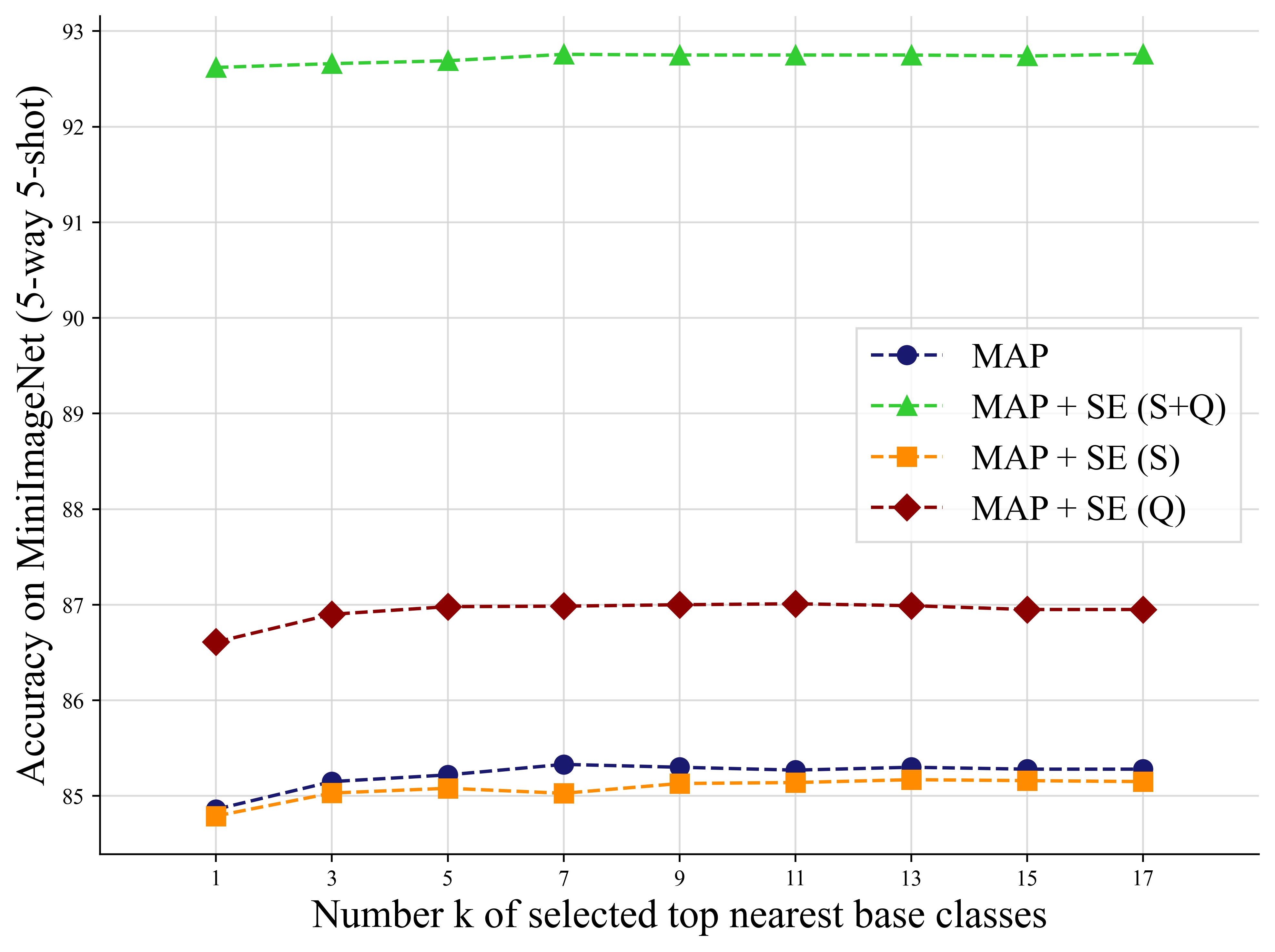}
		\caption{}
		\label{fig:topk_5shot}
	\end{subfigure}
	
\caption{
	Sensitivity of PMCE to the prior weight $\alpha$ and the number of retrieved base classes $k$ on MiniImageNet.
	Performance is stable over a wide range of values, and our default settings lie in the high-accuracy region.
}

	\label{fig:parameter_sensitivity}
\end{figure}

% heatmap
\begin{figure}[t]
	\centering
	% 如果你转成 pdf/eps，就把后缀改一下
	\includegraphics[width=0.8\columnwidth]{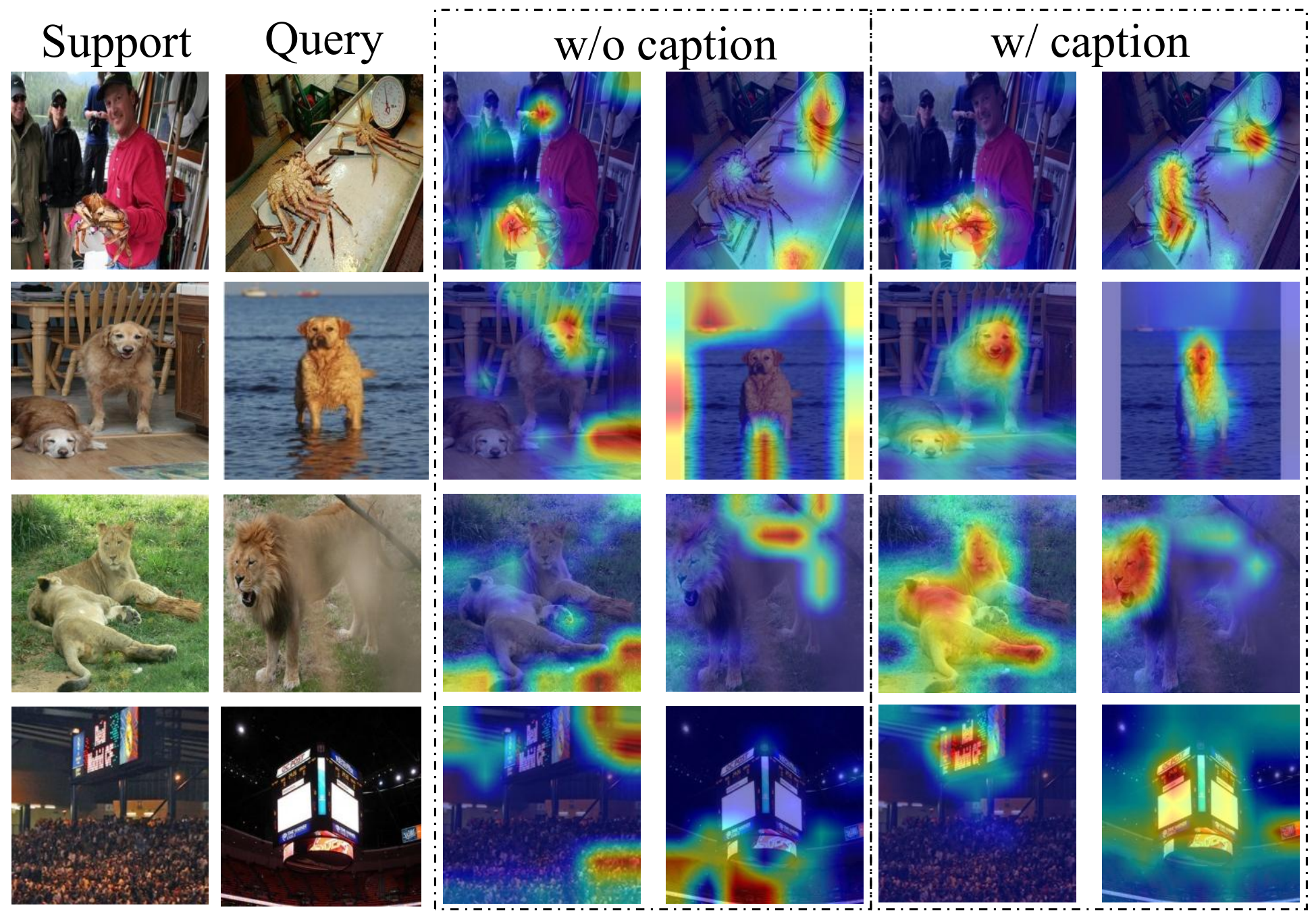}
\caption{Grad-CAM visualizations on the MiniImageNet test set. Compared with \textbf{w/o caption}, \textbf{w/ caption} produces more focused activations on query images, suppressing background responses and highlighting discriminative object regions.}
\label{fig:heatmap}
\end{figure}

%
%\begin{figure}[!htb]
%	\centering
%	\begin{minipage}{0.48\linewidth}
%		\centering
%		\includegraphics[width=\linewidth]{figs/SemanticPrior_LR_1shot_ClassName_vs_CaptionGuided_20251224_110723.png}
%		\vspace{-0.5em}
%		{\small (a) 5-way 1-shot}
%	\end{minipage}
%	\hfill
%	\begin{minipage}{0.48\linewidth}
%		\centering
%		\includegraphics[width=\linewidth]{figs/SemanticPrior_LR_5shot_ClassName_vs_CaptionGuided_20251224_110723.png}
%		\vspace{-0.5em}
%		{\small (b) 5-way 5-shot}
%	\end{minipage}
%\caption{
%	Comparison of semantic distance cues for MAP prior selection on MiniImageNet. We contrast class-name semantics and caption-guided semantics in 5-way 1-shot (a) and 5-way 5-shot (b). Class-name semantics yield more reliable retrieval in the extreme 1-shot case, while the gap becomes small in the 5-shot setting.
%}
%
%	\label{fig:map_semantic_distance}
%\end{figure}

% 设置边框的粗细（可选，0.4pt为默认细边框）
\setlength{\fboxsep}{0pt}
\setlength{\fboxrule}{0.4pt}

\begin{figure}[!htb]
	\centering
	% 整体缩放比例（例如 0.85）
	\scalebox{0.9}{
		\begin{minipage}{0.88\columnwidth}
			\centering
			\captionsetup[subfigure]{font=small}
			
			% 第一行两个
			\begin{subfigure}[b]{0.45\columnwidth}
				\centering
				\fbox{\includegraphics[width=\textwidth]{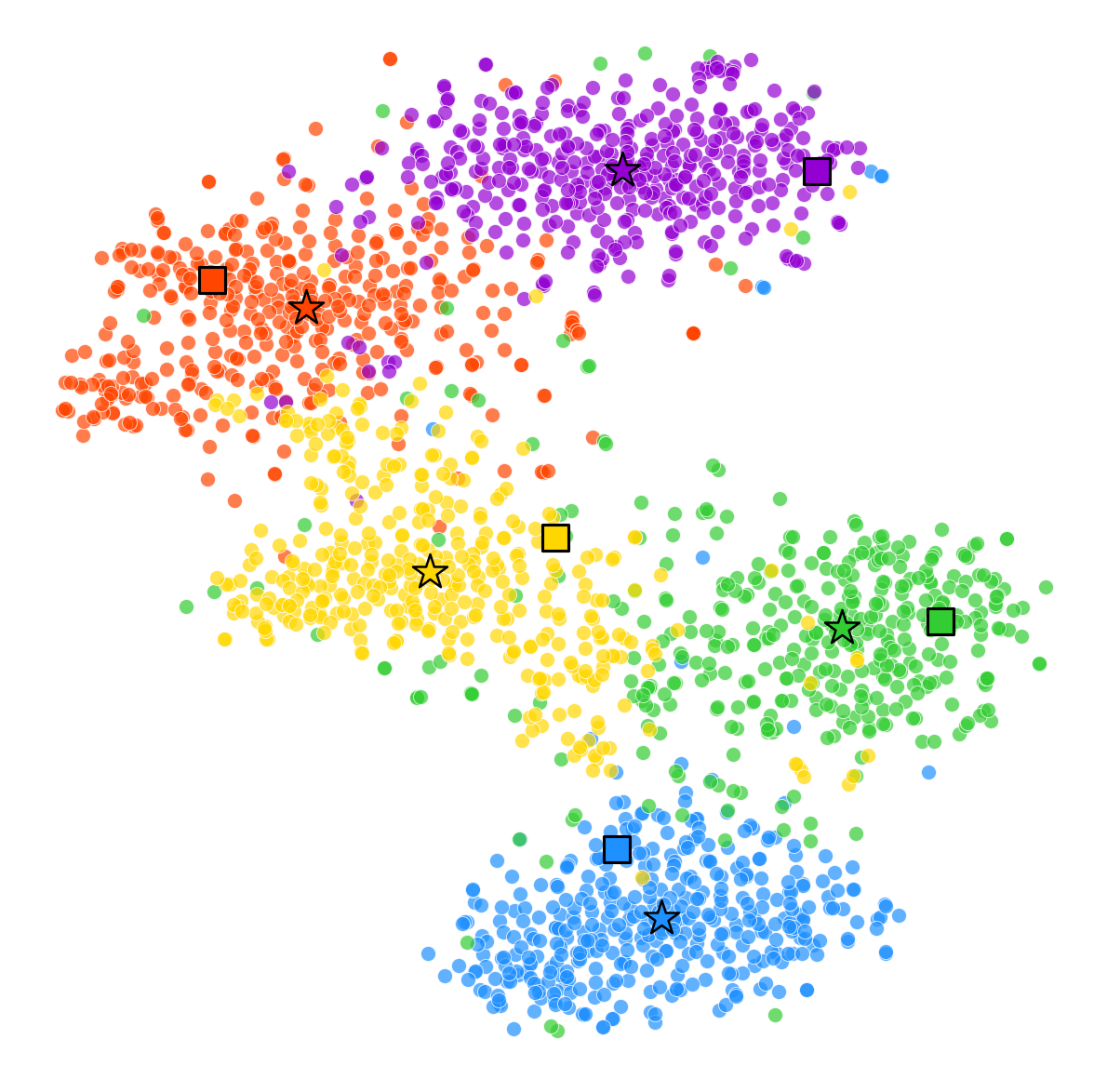}}
				\caption{Initial Prototype}
				\label{fig:init_proto}
			\end{subfigure}
			\hfill
			\begin{subfigure}[b]{0.45\columnwidth}
				\centering
				\fbox{\includegraphics[width=\textwidth]{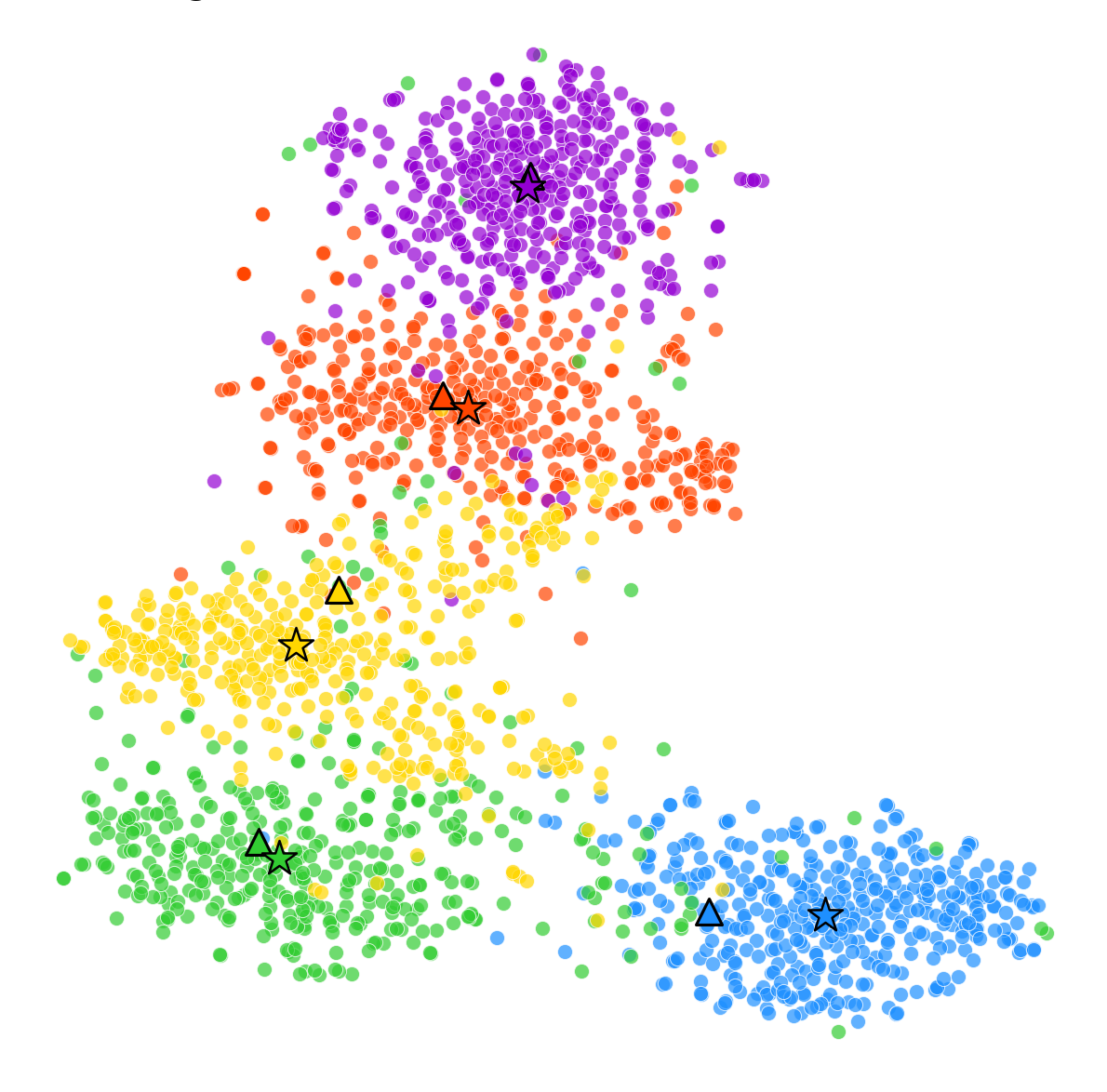}}
				\caption{MAP Prototype}
				\label{fig:map_proto}
			\end{subfigure}
			
			\vspace{8pt}
			
			% 第二行两个
			\begin{subfigure}[b]{0.45\columnwidth}
				\centering
				\fbox{\includegraphics[width=\textwidth]{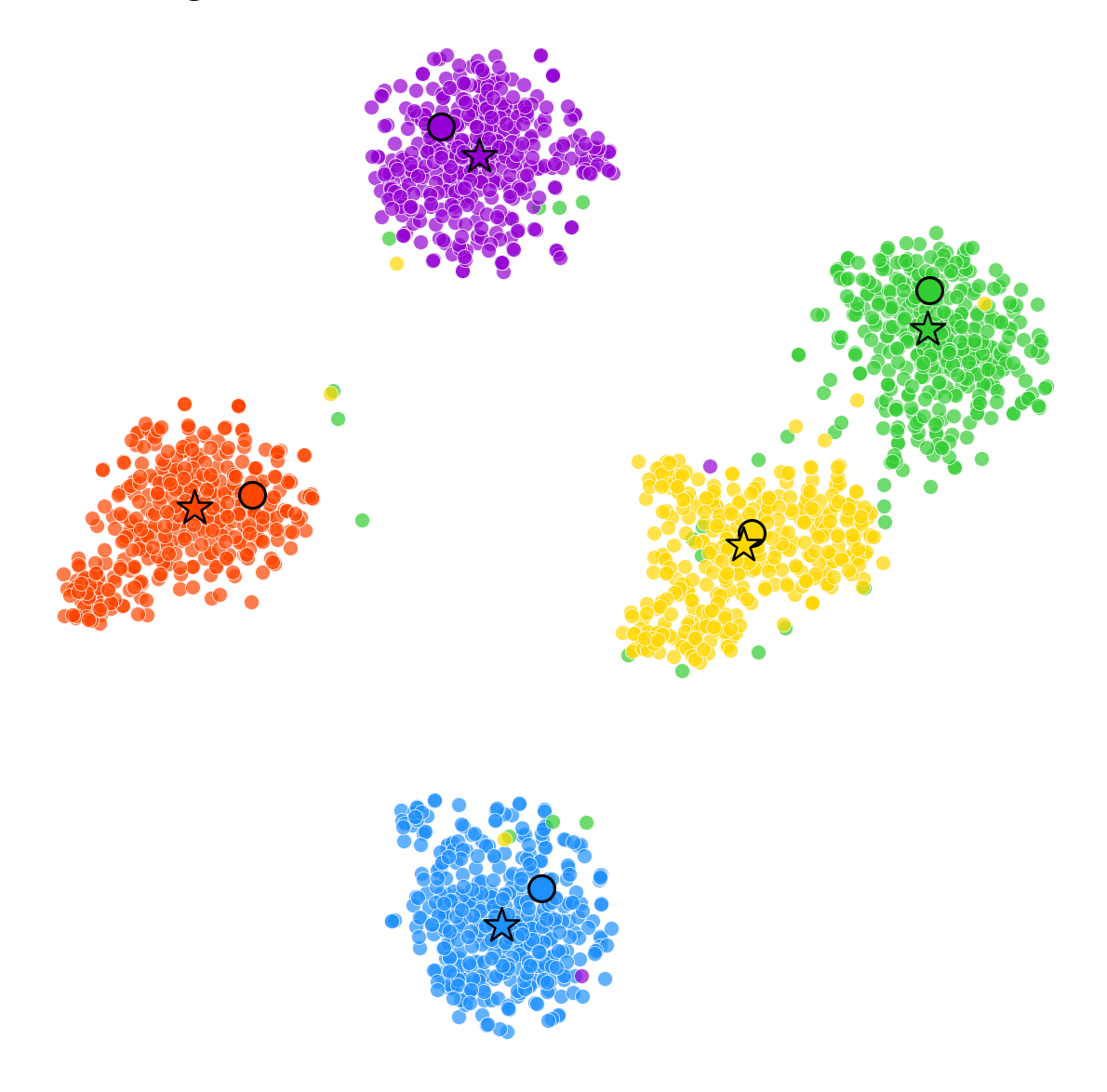}}
				\caption{Enhanced Prototype}
				\label{fig:enh_proto}
			\end{subfigure}
			\hfill
			\begin{subfigure}[b]{0.45\columnwidth}
				\centering
				\fbox{\includegraphics[width=\textwidth]{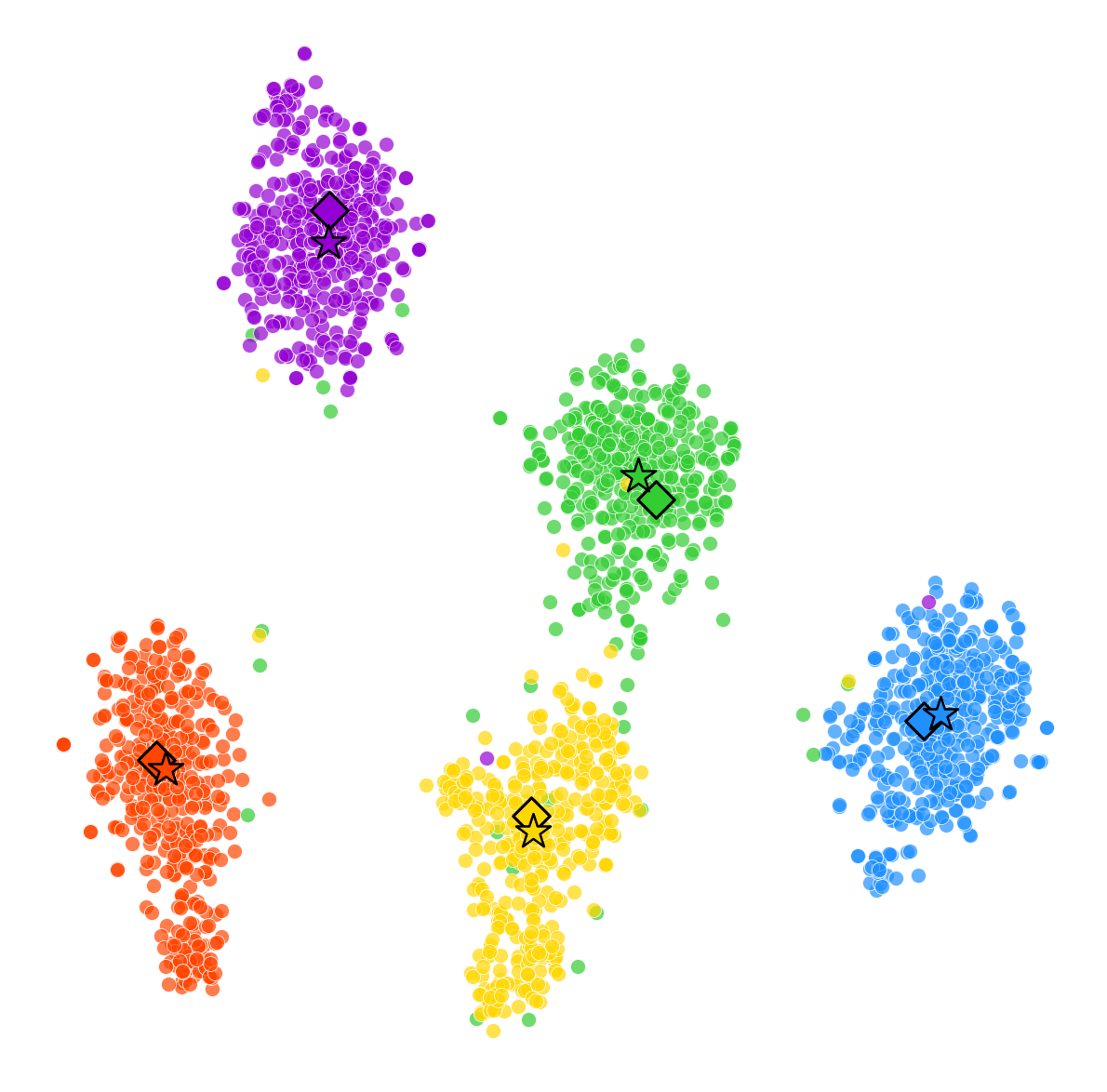}}
				\caption{Final Prototype}
				\label{fig:final_proto}
			\end{subfigure}
			
		\end{minipage}
	} % scalebox 结束
	
\caption{
	t-SNE visualization of feature distributions and prototype evolution on a 5-way 1-shot task from MiniImageNet.
	Different colors denote classes; $\star$ are ground-truth centers, $\blacksquare$ initial prototypes, $\triangle$ MAP rectified prototypes,large $\bullet$ caption-enhanced prototypes, and $\diamond$ the final prototypes of PMCE.
	MAP calibration and caption-guided enhancement jointly move prototypes toward the true centers and make clusters more compact.
}

	\label{fig:tsne_comparison}
\end{figure}

\subsection{Experimental Results}

We compare PMCE with representative metric-based methods FGFL~\cite{cheng2023frequency} and CPEA~\cite{hao2023cpea}, distribution-based methods MAP~\cite{wu2021feature} and ImprovedMAP~\cite{wu2022improved}, and recent semantic and vision--language baselines BMI~\cite{DBLP:conf/mm/LiW23}, SP-CLIP~\cite{DBLP:journals/corr/abs-2303-14123}, SemFew-Trans~\cite{zhang2024semfew}, and PRSN~\cite{dong2025prsn}.
Results on MiniImageNet and TieredImageNet are reported in Tab.~\ref{tab:compare_mini_tiered}, and results on CIFAR-FS and FC100 are reported in Tab.~\ref{tab:compare_cifar_fc100}.

Across ImageNet-style benchmarks, PMCE consistently improves over strong semantic and vision--language baselines under both ResNet-12 and Swin-T. 
The gains are most apparent in the 1-shot regime, where prototype bias is severe and query-side refinement becomes critical. 
For example, as is shown in Tab.~\ref{tab:compare_mini_tiered}, with Swin-T, PMCE reaches $85.03\%$ on MiniImageNet 1-shot and surpasses SemFew-Trans~\cite{zhang2024semfew} by  $7.71\%$ in this setting. 
Similar improvements are observed on TieredImageNet, indicating that the benefit transfers to a larger and more diverse label space.

On CIFAR-100 derived benchmarks, the $32\times32$ resolution makes captions less specific, which naturally attenuates the advantage of instance-level semantics. Nevertheless, as shown in Tab.~\ref{tab:compare_cifar_fc100}, PMCE remains competitive on CIFAR-FS and achieves the best 1-shot accuracy of 80.92\%, surpassing PRSN by 2.79\%. On FC100, PMCE achieves the top performance with 52.46\% and 67.00\%, yielding improvements over PRSN by 9.5\% and 6.43\%, respectively.

Overall, PMCE delivers consistent gains across backbones, with the most stable improvements on ImageNet-style benchmarks where captions are more informative. We believe the improvements mainly come from two design choices. First, semantics-guided prior retrieval provides a class-specific prior that stabilizes prototype estimation when supports are extremely limited, which is especially important in 1-shot. Second, PMCE optimizes the query side with label-free captions under the inductive protocol, so support and query features are better aligned before matching.

\subsection{Empirical Analysis}

\textbf{Ablation study.}
Tab.~\ref{tab:ablation_7rows} analyzes key components under the 5-way 1-shot setting on MiniImageNet and FC100 with the Swin-T backbone.
The baseline achieves $71.02\%$ on MiniImageNet and $45.27\%$ on FC100.
Adding semantics-guided MAP calibration alone improves MiniImageNet to $74.93\%$, which corresponds to a gain of $5.51\%$.
Applying caption-guided enhancement to a single stream yields smaller gains, and query-side enhancement is more influential than support-side enhancement.
Enhancing both streams without MAP reaches $79.42\%$ on MiniImageNet and $48.79\%$ on FC100.
Combining MAP calibration with dual-stream enhancement yields the best performance, reaching $85.03\%$ on MiniImageNet and $52.46\%$ on FC100, which improves over the baseline by $19.73\%$ and $15.88\%$.

\textbf{Impact of hyperparameters.}
Fig.~\ref{fig:parameter_sensitivity} explores the prior weight $\alpha$ in Eq.~(\ref{eq:map_pre}) and the neighbor count $k$ assigned for prior retrieval.
In 1-shot, the high-accuracy region concentrates around $\alpha$ in the range of 0.3 to 0.4.
In 5-shot, the best region shifts toward a larger $\alpha$ around 0.7.
Accuracy remains stable across a wide range of $k$, indicating that the retrieval step is not critically dependent on the precise count of base classes.

\subsection{Visualization Analysis}

\textbf{Effect of caption guidance.}
Fig.~\ref{fig:heatmap} compares Grad-CAM activation maps with and without caption guidance.
Without captions, query responses are often diffuse and may include background regions.
With caption guidance, activations become more concentrated on object regions and less affected by background clutter.

\textbf{Prototype evolution.}
Fig.~\ref{fig:tsne_comparison} visualizes feature distributions and prototype evolution on a 5-way 1-shot episode from MiniImageNet.
MAP calibration shifts prototypes toward more reasonable regions by injecting class-specific priors, and caption-guided enhancement further tightens clusters and improves inter-class separation.
The final prototypes are closer to the ground-truth centers, supporting that PMCE improves prototype estimation and query-side representations in a unified metric space.

\section{Conclusion}

In this paper, we have presented PMCE, a probabilistic few-shot framework that integrates semantics-guided prior selection, MAP prototype calibration, and caption-guided feature optimization. By using class-name embeddings to select class-specific priors from base statistics and leveraging BLIP captions as label-free instance cues for both supports and queries, PMCE produces more stable prototypes and better-aligned query representations, with a lightweight consistency regularizer to mitigate caption noise. Experiments on four benchmarks with two backbones show consistent improvements over strong metric-, semantic-, and distribution-based baselines, with the most pronounced gains in the 1-shot setting. Future work will explore the confidence of captions to further reduce semantic drift under noisy semantics.

\cleardoublepage
\appendix
\section{Appendix}
\subsection{Additional Implementation Details for PMCE}

To ensure the reproducibility of our PMCE framework, we provide the detailed algorithmic procedures for the two core components: Semantics-Guided Prior Selection and Caption-Guided Feature Enhancement.

\subsubsection{Semantics-guided MAP Prototype Calibration} 
Algorithm~\ref{alg:prototype_calibration} details how we rectify support prototypes by leveraging priors from the base classes.

\textbf{Algorithmic Steps.}Our procedure begins by retrieving the top-$K$ semantically similar base classes. We then aggregate their visual means to form $\mu_{prior}$.This prior is then fused with the support prototype $p_{init}$ to compute the calibrated prototype $\tilde{p}$. The hyperparameter $\alpha$ is employed to balance the contribution between the support information and the retrieved base-class knowledge across different shot settings.

\begin{algorithm}
\caption{Semantics-guided MAP Prototype Calibration (Weighted Prior)}
\label{alg:map_weighted}
\small
\textbf{Input:} Support set $\mathcal{S}$, novel class semantic $s_{\text{class}}$, Knowledge Bank $\mathcal{B}=\{(\mu_j, s_{\text{base}}^j)\}_{j=1}^{|\mathcal{C}_b|}$ \\
\textbf{Parameter:} Number of neighbors $K$, balancing hyperparameter $\alpha$, temperature $\tau$ \\
\textbf{Output:} Rectified prototype $\tilde{p}$
\begin{algorithmic}[1]
\State Compute the empirical prototype $p_{\text{init}}$ by averaging features of $\mathcal{S}$.
\For{each base class $(\mu_j, s_{\text{base}}^j)\in\mathcal{B}$}
    \State Calculate semantic similarity $\text{score}_j=\cos(s_{\text{class}}, s_{\text{base}}^j)$ according to Eq.~(3).
\EndFor
\State Select Top-$K$ indices to form the neighbor set $\mathcal{N}_{\text{top}}$.
\State Compute normalized weights $w_k=\frac{\exp(\text{score}_k/\tau)}{\sum_{u\in\mathcal{N}_{\text{top}}}\exp(\text{score}_u/\tau)}$, $\forall k\in\mathcal{N}_{\text{top}}$.
\State Aggregate visual priors: $\mu_{\text{prior}}=\sum_{k\in\mathcal{N}_{\text{top}}} w_k\,\mu_k$.
\State Calculate $\tilde{p}=\alpha p_{\text{init}}+(1-\alpha)\mu_{\text{prior}}$ following Eq.~(6).
\State \Return $\tilde{p}$
\end{algorithmic}
\label{alg:prototype_calibration}
\end{algorithm}

\begin{algorithm}
\caption{Forward Pass of the Caption-Guided Enhancer $\Phi$ }
\label{alg:enhancer_notoken}
\small
\textbf{Input:} Visual feature $v_{\text{in}}\in\mathbb{R}^{1\times d_v}$, caption embedding $\mathbf{S}_{\text{in}}\in\mathbb{R}^{d_t}$ \\
\textbf{Parameter:} Projection layer $(W_p,b_p,\text{LN},\text{ReLU})$, multi-head cross-attention $\text{Attn}$ (heads $h$), projections $(W_Q,W_K,W_V)$, scaling factor $\beta$ \\
\textbf{Output:} Enhanced feature $v_{\text{out}}$
\begin{algorithmic}[1]
\State \textbf{Semantic Projection (token-wise):}
$\mathbf{S}_{\text{proj}}\leftarrow \text{ReLU}(\text{LN}(\mathbf{S}_{\text{in}}W_p+b_p))\in\mathbb{R}^{d_v}$.
\State \textbf{Cross-Attention (visual token $\rightarrow$ caption tokens):}
$\mathbf{Q}\leftarrow v_{\text{in}}W_Q\in\mathbb{R}^{1\times d_k}$,
$\mathbf{K}\leftarrow \mathbf{S}_{\text{proj}}W_K\in\mathbb{R}^{1\times d_k}$,
$\mathbf{V}\leftarrow \mathbf{S}_{\text{proj}}W_V\in\mathbb{R}^{1\times d_k}$.
\State Compute attention map:
$\mathbf{A}\leftarrow \text{Softmax}\!\big(\frac{\mathbf{Q}\mathbf{K}^\top}{\sqrt{d_k}}\big)\in\mathbb{R}^{1\times T}$.
\State Obtain attended semantic residual: $\Delta v \leftarrow \mathbf{A}\mathbf{V}\in\mathbb{R}^{1\times d_k}$.
\State (Multi-head) Apply the above per head and concatenate/project to match $d_v$ (standard MHA).
\State \textbf{Residual Integration:} $v_{\text{out}}\leftarrow v_{\text{in}}+\beta\cdot \Delta v$.
\State \Return $v_{\text{out}}$
\end{algorithmic}
\label{alg:enhancer}
\end{algorithm}

\subsubsection{Details of the Caption-Guided Enhancer Module}

\begin{figure}
    \centering
    \includegraphics[width=\columnwidth]{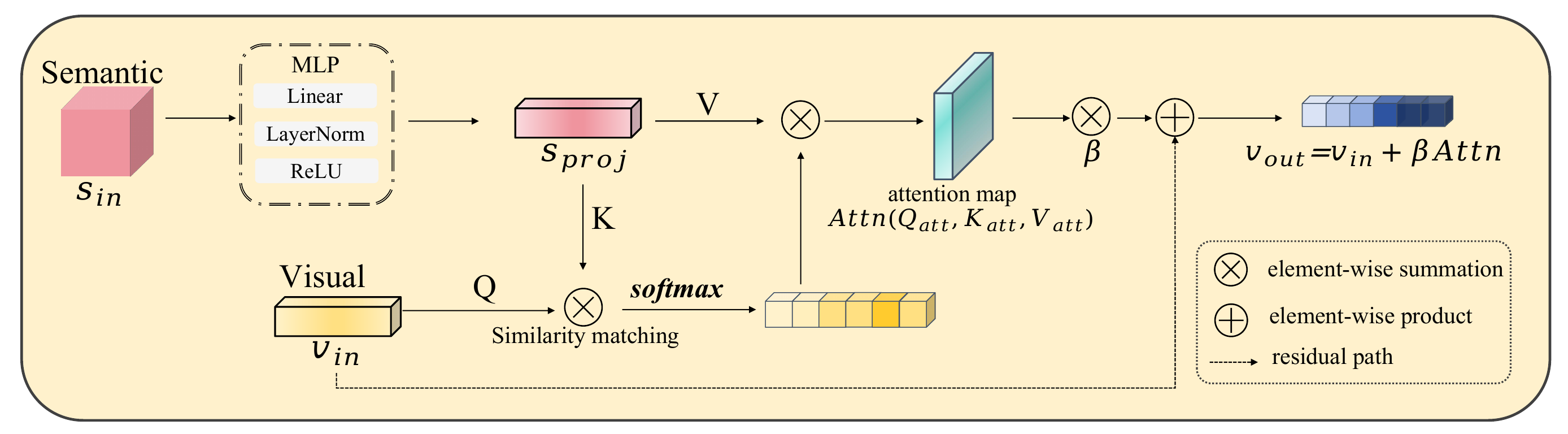}
    \caption{
        Architecture of the caption-guided enhancer $\Phi$.
        Instance-level semantics from captions are projected and injected
        into visual features via cross-attention, and a learnable
        factor $\beta$ controls the strength of the semantic residual.
    }
    \label{fig:enhancer_detail}
\end{figure}

To supplement the methodology in Section~\ref{sec:enhancement}, we provide further implementation details of the Caption-Guided Enhancer $\Phi$.Algorithm~\ref{alg:enhancer} details the overall execution logic for this process.As illustrated in Figure~\ref{fig:enhancer_detail}, $\Phi$ is designed as a lightweight cross-modal residual block that refines visual representations by leveraging instance-level semantic contexts.

\textbf{Feature Interaction.} Consistent with Eqs.~(\ref{eq:semantic_projection})--(\ref{eq:vout}), the module first aligns the caption embedding $s_{in}$ to the visual space via a projection layer: $s_{proj} = \text{ReLU}(\text{LN}(W_p s_{in} + b_p))$. We then model the primary interaction through a multi-head cross-attention mechanism, configuring the number of heads to $h = 4$. Utilizing $h$ attention heads enables the model to capture diverse information from multiple semantic subspaces simultaneously.

\textbf{Gated Residual Connection.} To keep the pre-trained backbone's discriminative power intact, we integrate the attention output through a gated residual path : $v_{out} = v_{in} + \beta Attn(Q, K, V)$, following the formulation in Eq.\ref{eq:vout}. Notably, the learnable scaling factor $\beta$ is initialized to 0.1. Such an initialization strategy provides an initial momentum for visual-semantic fusion. More importantly, it shields the model from overlooking semantic cues or experiencing unstable gradients as training begins.

\subsection{Impact of Classification Strategies and Retrieval Cues}
\begin{figure}[t]
    \centering
    \begin{subfigure}[t]{0.48\linewidth}
        \centering
        \includegraphics[width=\linewidth]{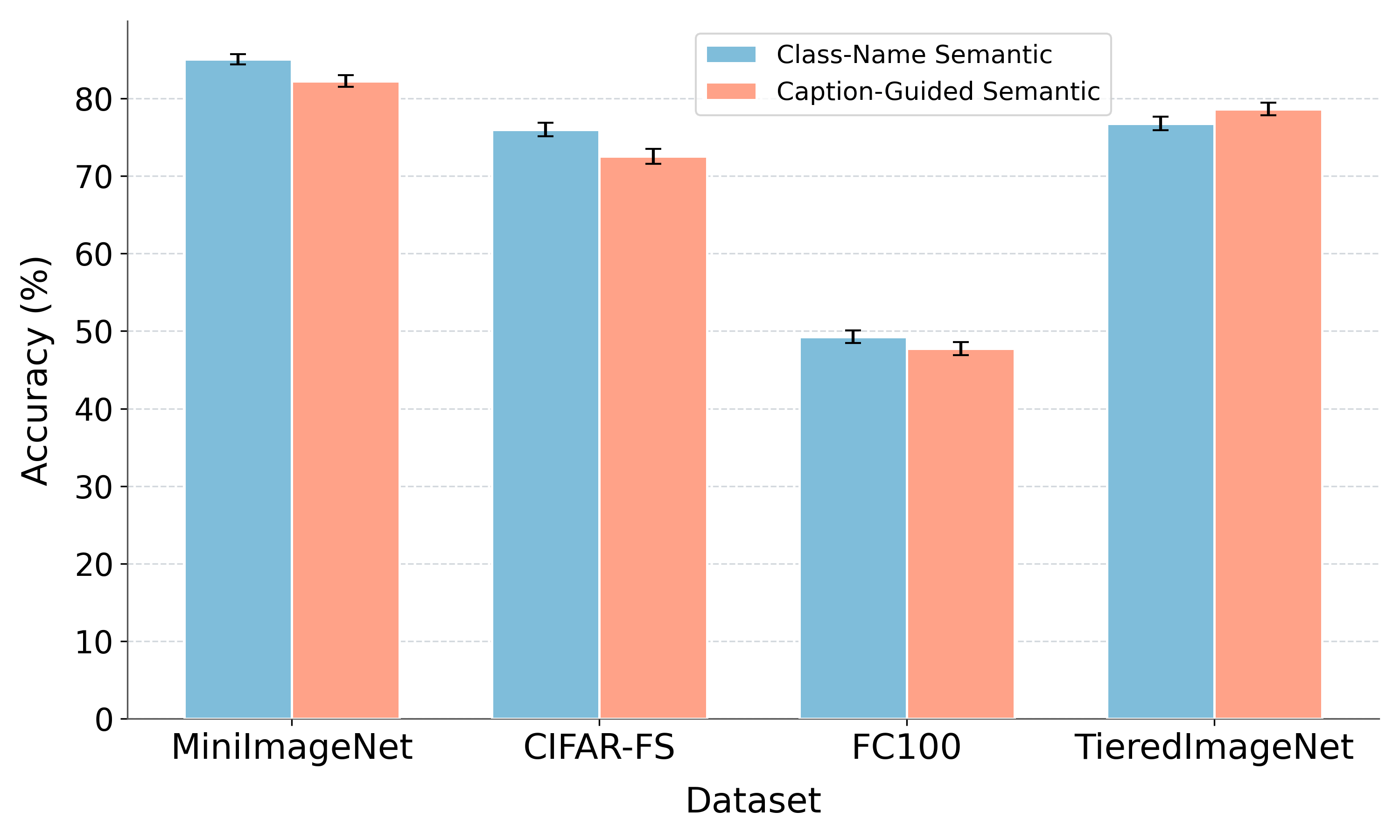}
        \subcaption{5-way 1-shot}
        \label{fig:map_semantic_distance_1shot}
    \end{subfigure}
    \hfill
    \begin{subfigure}[t]{0.48\linewidth}
        \centering
        \includegraphics[width=\linewidth]{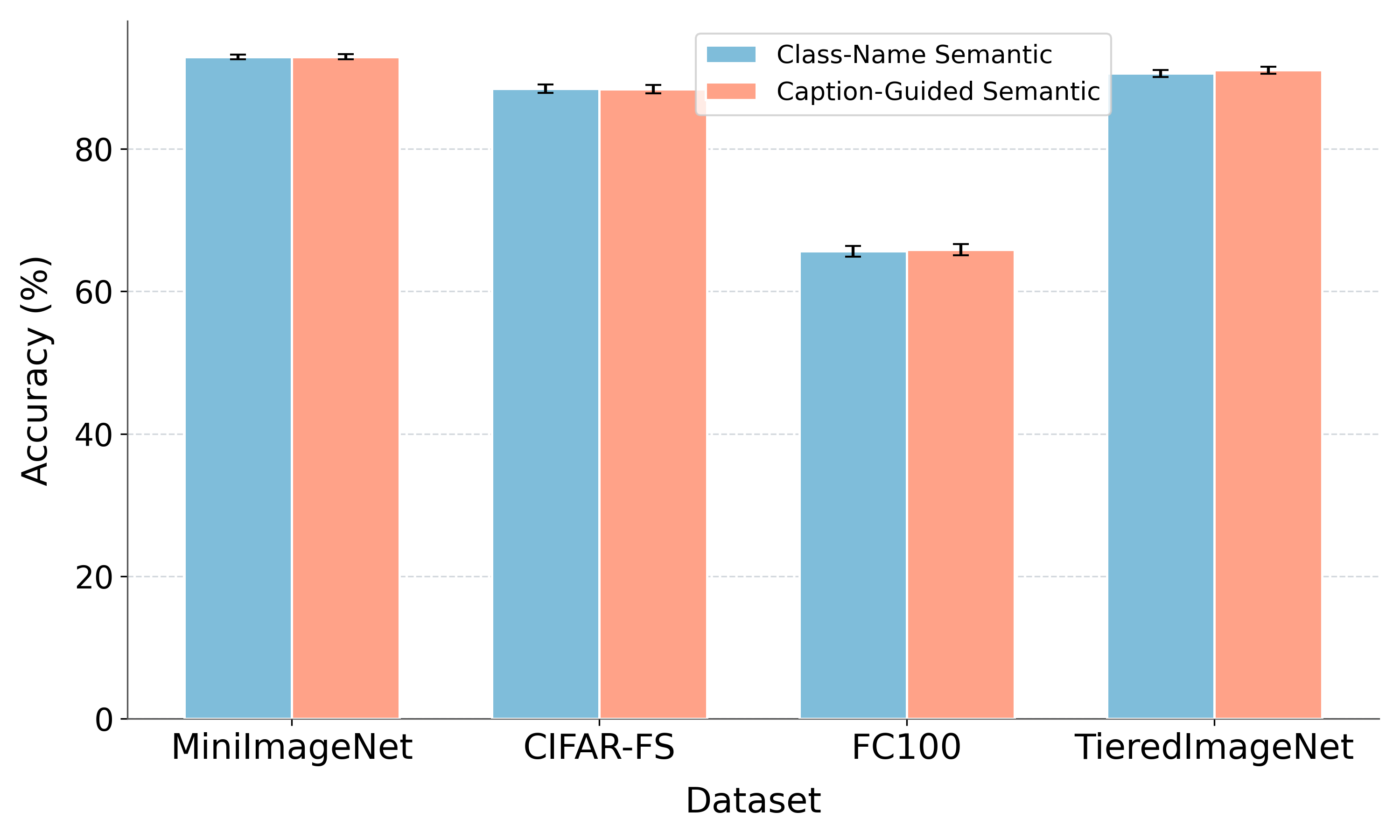}
        \subcaption{5-way 5-shot}
        \label{fig:map_semantic_distance_5shot}
    \end{subfigure}

    \caption{
    Comparison of different semantic distance cues used for distance-based prior candidate selection in the MAP framework.
    The semantic information here is used to compute distances for selecting relevant base-class statistics in MAP,
    rather than being directly modeled as a probabilistic prior or used for feature enhancement.
    }
    \label{fig:map_semantic_distance}
\end{figure}

Figure~\ref{fig:map_semantic_distance} illustrates a comparison between two distinct retrieval strategies.Initially, we use coarse-grained class-name semantics for both the base and support sets.The second uses concatenated class-name and caption semantics for the base set and the averaged caption features of the support samples for the current task.As shown in Fig.~\ref{fig:map_semantic_distance_1shot}, class-name semantics provide more accurate retrieval in the 1-shot setting. In contrast, under the 5-shot setting (Fig.~\ref{fig:map_semantic_distance_5shot}), the difference between the two strategies becomes negligible.Due to its performance in extreme low-data regimes, we adopt the class-name semantics as our default retrieval cue.We further assess the effect of the classifier on the final accuracy.Table~\ref{tab:classifier_comparison} presents the results of three classifiers,in which logistic classifier achieves the best overall performance,the margins between the three methods are remarkably narrow.

% ==========================================================
% map_semantic_distance figure
% ==========================================================

% ==========================================================
% Classifier comparison table
% ==========================================================
\begin{table}[H]
\centering
\caption{Average classification accuracy (\%) of different classifiers under the same MAP framework.
All results are reported using a fixed semantic distance cue for distance-based prior candidate selection in MAP.
LR denotes the linear logistic regression classifier, EU denotes the Euclidean-distance nearest-prototype classifier,
and CO denotes the cosine-distance nearest-prototype classifier.}
\footnotesize
\resizebox{\columnwidth}{!}{%
\begin{tabular}{c cccccccc}
\toprule
\textbf{Classifier} &
\multicolumn{2}{c}{\textbf{MiniImageNet}} &
\multicolumn{2}{c}{\textbf{TieredImageNet}} &
\multicolumn{2}{c}{\textbf{CIFAR-FS}} &
\multicolumn{2}{c}{\textbf{FC100}} \\
\cmidrule(lr){2-3} \cmidrule(lr){4-5} \cmidrule(lr){6-7} \cmidrule(lr){8-9}
& \textbf{1-shot} & \textbf{5-shot}
& \textbf{1-shot} & \textbf{5-shot}
& \textbf{1-shot} & \textbf{5-shot}
& \textbf{1-shot} & \textbf{5-shot} \\
\midrule
EU & 83.84 & 92.59 & 82.04 & 93.07 & 79.44 & 88.30 & 52.01 & 67.16 \\
CO & 83.83 & 92.60 & 82.91 & 93.08 & 79.48 & 88.19 & 52.28 & 66.98 \\
LR & 85.03 & 92.77 & 83.13 & 93.23 & 80.92 & 89.02 & 52.46 & 67.00 \\
\bottomrule
\end{tabular}
} % end resizebox

\label{tab:classifier_comparison}
\end{table}

% ==========================================================
% cross-domain table
% ==========================================================
\begin{table}[H]
\centering
\caption{The 5-way 1-shot and 5-way 5-shot accuracy (\%) for cross-domain on MiniImageNet $\rightarrow$ CUB. The best results are highlighted in bold.}
\footnotesize
\resizebox{\columnwidth}{!}{%
\begin{tabular}{l c c c} % 第一列左对齐，其余居中
\toprule
% 表头第一行
% Method 和 Semantic 纵向跨2行 (*) 表示自动宽度
\multirow{2}{*}{\textbf{Method}} & \multirow{2}{*}{\textbf{Semantic}} & \multicolumn{2}{c}{\textbf{MiniImageNet} $\rightarrow$ \textbf{CUB}} \\
\cmidrule(lr){3-4} % 横线只画在第3-4列下面
% 表头第二行 (Method 和 Semantic 位置留空，因为被上面占用了)
 & & \textbf{5-way 1-shot} & \textbf{5-way 5-shot} \\
\midrule
% 数据行 - Semantic 列已清空，1-shot 列用 --- 占位
MatchNet~\cite{vinyals2016matching} & No & 42.62 ± 0.55 &  56.53 ± 0.44 \\
ProtoNet~\cite{snell2017prototypical} & No & 50.51 ± 0.56  & 69.28 ± 0.40 \\
MAML~\cite{finn2017model}  & No & 43.59 ± 0.54  & 54.18 ± 0.41 \\
RelationNet~\cite{sung2018learning}       & No & 49.84 ± 0.54  & 68.98 ± 0.42 \\
SimpleShot~\cite{wang2019simpleshot}        & No & 48.56 & 65.63\\
GNN + FT~\cite{DBLP:conf/iclr/TsengLH020}      & No & 47.47 ± 0.75  & 66.98 ± 0.68 \\
MN + AFA ~\cite{hu2022adversarial}      & No & 41.02 & 59.46 \\
StyleAdv~\cite{fu2023styleadv} & No & 48.49 & 68.72 \\
LDP-net ~\cite{zhou2023revisiting} & No & 49.82 & 70.39 \\
MIFN ~\cite{zhou2023revisiting} & Yes & 48.21 ± 0.60  & 65.33 ± 0.54 \\
MIFN+Ltri ~\cite{zhou2023revisiting} & Yes & 49.21 ± 0.60 & 66.30 ± 0.56 \\
FLoR ~\cite{zou2024flatten} & No & 49.99 & 70.39 \\
TPN-FAP ~\cite{Zhang2024Exploring} & Yes & 50.56 ± 0.50 &  64.17 ± 0.40 \\
TA-MLA ~\cite{yue2025multi} & Yes & 48.57 ± 0.47 &  69.92 ± 0.45 \\

\midrule
\textbf{PMCE (ours)}   & Yes & \textbf{52.09 $\pm$ 0.88} & \textbf{70.79 $\pm$ 0.78} \\
\bottomrule
\end{tabular}
}
\label{tab:cross_domain_cub_semantic}
\end{table}

\subsection{Evaluation on Cross-Domain Adaptation}

To assess the cross-domain generalization of PMCE, we further evaluate it in the cross-domain few-shot setting, where episodes are sampled from a target dataset with a distribution different from the base training data. The quantitative results are reported in Tab.~\ref{tab:cross_domain_cub_semantic}. Overall, PMCE attains accuracy comparable to recent strong baselines under this challenging protocol. In the 5-way 5-shot case, it slightly surpasses the metric-based LDP-Net by 0.40\%, while remaining on par with it in the 1-shot setting.

These results indicate that the proposed caption-guided enhancement can still provide useful auxiliary cues when there is a noticeable domain gap, although the absolute accuracies are naturally lower than in the in-domain scenario. Semantic cross-attention thus offers a modest but consistent benefit for cross-domain generalization, especially when a few more labeled examples are available.

\begin{figure*}[t]
    \centering
    % trim=左 下 右 上 (单位可以是 cm, mm, 或 pt)
    % clip 参数必须加上，否则裁剪不会生效
    \includegraphics[width=0.75\textwidth, trim=0cm 0cm 0cm 0cm, clip]{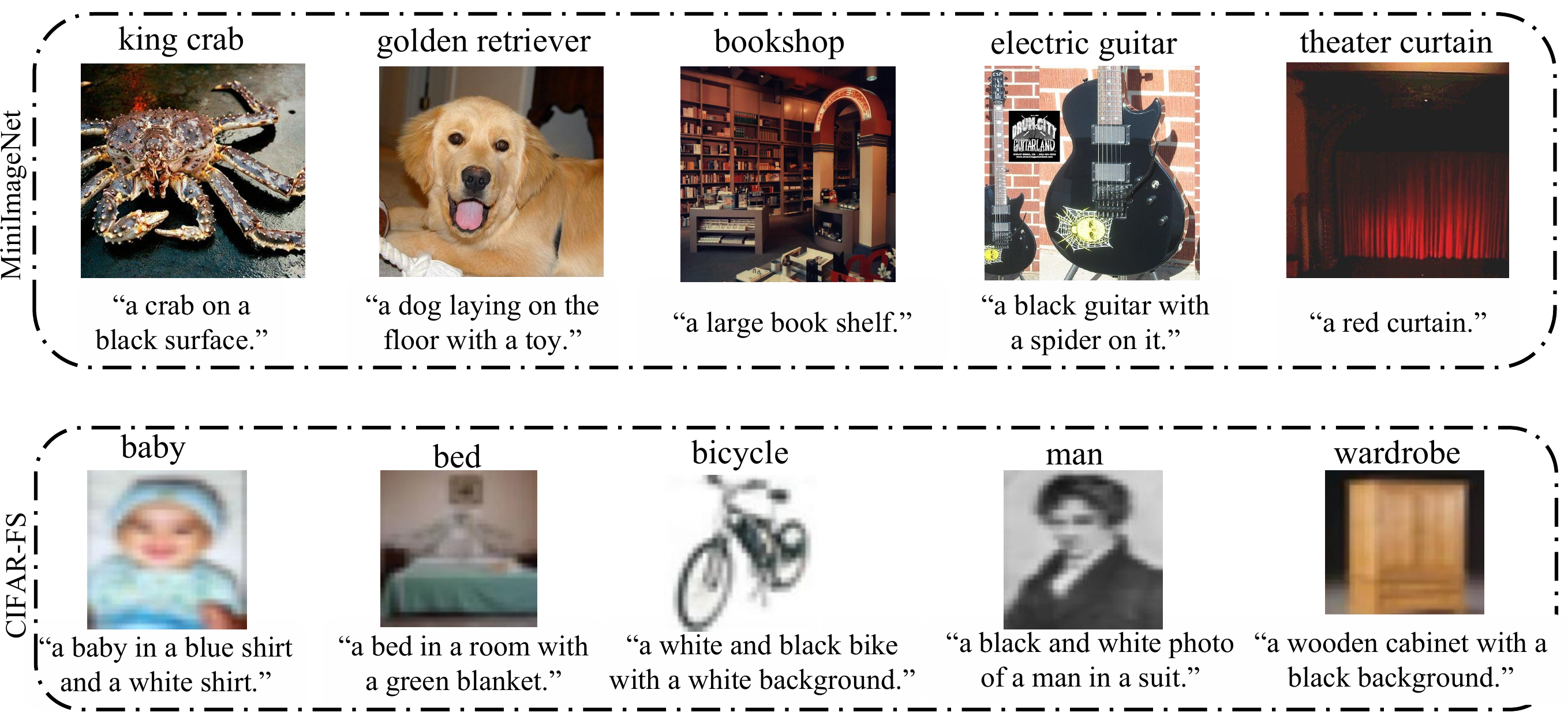} 
    \caption{Visualizations of example images from MiniImageNet and CIFAR-FS datasets. For each sample, the category name is displayed above the image, while the generated descriptive caption is provided below.}
    \label{fig:caption_example}
\end{figure*}

\subsection{BLIP Pre-training and Potential Data Overlap}

We use the \textit{blip-image-captioning-base} model to generate instance-level captions. This captioner is pre-trained for image–text generation on a mixture of human-annotated datasets (COCO~\cite{lin2014microsoft}, Visual Genome~\cite{krishna2017visual}) and web-scale collections (Conceptual Captions and Conceptual 12M~\cite{changpinyo2021conceptual}, SBU captions~\cite{ordonez2011im2text}, LAION~\cite{DBLP:journals/corr/abs-2111-02114}). In our few-shot pipeline, BLIP is kept frozen and is only applied to images from the base and novel sets, similar to a fixed visual backbone. The model is trained for caption generation rather than classification, and we do not fine-tune it on any of the few-shot benchmarks (MiniImageNet, TieredImageNet, CIFAR-FS, FC100). Although some overlap between web-scale pre-training corpora and benchmark images cannot be completely ruled out, any such overlap is indirect and does not reuse class labels. Conceptually, this setting is analogous to using an ImageNet- or CLIP-pretrained encoder as a frozen feature extractor, and the improvements of PMCE mainly come from how captions are used as label-free semantic cues within an inductive protocol rather than from task-specific pre-training on the target datasets.

In Figure~\ref{fig:caption_example}, we present several examples from MiniImageNet and CIFAR-FS. Here, ground-truth labels appear above the images while the corresponding BLIP captions are listed below.Instead of reproducing fine-grained class names, the generated captions focus on visible attributes like color, pose, background, and coarse object categories. Taking the "golden retriever" class as an example, the model typically generates descriptions like "a dog lying on the floor with a toy." Similarly, it identifies a "king crab" simply as "a crab on a black surface".When handling low-resolution datasets such as CIFAR-FS, BLIP often yields brief, generic phrases. This behavior suggests that the output reflects inherent visual ambiguity rather than a direct retrieval of benchmark labels. Such results reinforce our interpretation of BLIP as a generic visual descriptor here, as opposed to a mere repository for memorized labels.

We kept the BLIP captioner frozen across all experimental phases, including both training and meta-test time. It is applied to support and query images in the same way and is never updated using support sets or query labels. Each query is processed independently, without access to the joint distribution of the query set. Under this setting, captions function as additional instance-level cues injected into an inductive few-shot pipeline, not as an extra supervised signal tuned on the target tasks.

\begin{table*}[t]
\centering
\caption{Qualitative comparison between PMCE and recent semantic-driven few-shot methods. We summarize the semantic source, signal granularity and interaction mechanism for each line of work.}
\label{tab:qualitative_comparison}
\footnotesize
\resizebox{\textwidth}{!}{%
\begin{tabular}{l l l l l}
\toprule
\textbf{Category} & \textbf{Methodology} & \textbf{Semantic Source} & \textbf{Signal Granularity} & \textbf{Interaction Mechanism} \\
\midrule
% Group A
\multirow{4}{*}{\shortstack[l]{Semantic-based\\(Class-level)}} 
& BMI~\cite{DBLP:conf/mm/LiW23} & Category Names (CLIP) & Coarse (Class-level) & Latent space alignment \\
& MAVSI~\cite{zhao2024multi} & Category Names (GloVe) & Coarse (Class-level) & Multi-attention interaction \\
& SEVPro~\cite{DBLP:conf/ijcai/CaiLH024} & Category Names (CLIP) & Coarse (Class-level) & Semantic-to-Visual Projection and Prototype Enhancement \\
& PRSN~\cite{dong2025prsn} & Category Names (BERT) & Coarse (Class-level) & Prototype Resynthesis with Cross-image Semantic Alignment \\
\midrule
% Group B
\multirow{6}{*}{\shortstack[l]{Vision-Language\\(Class-level)}} 
& LPE~\cite{Yang_2023_WACV} & Visual Attributes (VLM Embeddings) & Detailed (Class-centric) & Class-specific Part Filters and Latent Embedding \\
& SP-CLIP~\cite{DBLP:journals/corr/abs-2303-14123} & Category Names (LLM/CLIP) & Static (Class-centric) & Spatial and Channel-wise Prompt Tuning \\
& SemFew~\cite{zhang2024semfew} & LLM (Evolved descriptions) & Refined (Class-centric) & Semantic Alignment Network \\
& SYNTRANS~\cite{DBLP:conf/ijcai/TangHQ25} & LMMs (CLIP + LLM) & Synergistic (Class-centric) & Synergistic Knowledge Mining and Visual-Semantic Bridging \\
& SAFF~\cite{DBLP:conf/cvpr/Cumplido0R25} & VLM Semantics & Selective (Class-centric) & Slot-based Feature Filtering with Weighted Masking \\
& VT-FSL~\cite{DBLP:journals/corr/abs-2509-25033} & LLM-based Iterative Descriptions & Refined (Class-centric) & Cross-modal Geometric Alignment \\
\midrule
% Ours

\textbf{Ours} & \textbf{PMCE} & 
\begin{tabular}[t]{@{}l@{}} \textbf{Category Names (CLIP)} \\ \textbf{Instance Caption (BLIP)} \end{tabular} & 
\begin{tabular}[t]{@{}l@{}} \textbf{Coarse (Class-level)} \\ \textbf{Dynamic (Instance-level)} \end{tabular} & 
\begin{tabular}[t]{@{}l@{}} \textbf{Prior selection for MAP-based Calibration} \\ \textbf{Visual-to-semantic cross-attention for instance-level prototype alignment} \end{tabular} \\
\bottomrule
\end{tabular}
}
\end{table*}

\subsection{Qualitative Comparison with Semantic-Driven FSL}

To put PMCE in context, we briefly review recent semantic-driven few-shot methods from the last three years and compare them along three axes: where the semantic signal comes from, at which granularity it is defined, and how heavy the underlying models are and how they are trained. Table~\ref{tab:qualitative_comparison} summarizes representative approaches.

\textbf{Semantic-based methods with class-level labels.}
BMI~\cite{DBLP:conf/mm/LiW23}, MAVSI~\cite{zhao2024multi}, SEVPro~\cite{DBLP:conf/ijcai/CaiLH024} and PRSN~\cite{dong2025prsn} rely on category names or other coarse class-level descriptions as their only semantic source.
MAVSI encodes each class name with GloVe and injects these word vectors into the prototype space by multi-head attention.
BMI and SEVPro use CLIP text encoders to embed prompts such as ``a photo of a \{label\}'' and then adapt visual prototypes or metric heads, while PRSN explores linguistic structures in WordNet or BERT spaces and resynthesizes prototypes via cross-image semantic alignment.
Most of these methods attach relatively light semantic modules to a standard few-shot backbone and keep the overall model size comparable to classical metric-based pipelines.

In all cases, every image of a class shares the same semantic vector, which is static once the label is fixed.
This strengthens coarse category alignment but cannot capture fine-grained cues such as pose, background or local texture.
When classes are visually similar, for example on FC100, this class-invariant treatment limits how much additional discrimination the semantics can provide.

\textbf{Vision–language methods with class-level prompts.}
A second group uses larger language models or vision–language models to enrich class-level semantics before injecting them into the vision pipeline.
LPE~\cite{Yang_2023_WACV} predicts visual attributes with a VLM and uses them as class-specific part filters.
SP-CLIP~\cite{DBLP:journals/corr/abs-2303-14123} and SemFew~\cite{zhang2024semfew} let LLMs refine CLIP prompts into more informative textual descriptions and then tune CLIP with spatial or channel-wise prompts.
SYNTRANS~\cite{DBLP:conf/ijcai/TangHQ25} combines CLIP with an LLM to transfer semantic knowledge to a lighter vision encoder, while SAFF~\cite{DBLP:conf/cvpr/Cumplido0R25} uses VLM-derived semantics to drive slot-based feature filtering.
VT-FSL~\cite{DBLP:journals/corr/abs-2509-25033} iteratively generates class-centric descriptions and aligns them with visual features in a geometric manner.

Compared with the first group, these approaches obtain richer and often more detailed text for each category and sometimes use much heavier backbones or language decoders.
Several methods partially fine-tune CLIP or attach learnable adapters and prompt modules with millions of parameters.
This improves accuracy but also blurs the line between gains from semantic design and gains from scaling model capacity and task-specific tuning.
Moreover, the semantics are still essentially class-centric:
all images from “golden retriever” share one refined description, regardless of the scene or pose, so the semantic branch remains blind to the specific context of each query.

\textbf{Position of PMCE.}
PMCE differs from both groups in terms of semantic granularity and model usage.
First, class names are not used to directly reshape features or classifier weights.
CLIP text embeddings of category names act as keys for retrieving a small set of related base classes, and only their visual statistics enter the MAP-based prior for prototype calibration.
Semantics decide which priors are selected, while the priors themselves live in the visual space.

Second, PMCE introduces instance-level captions as a dynamic semantic branch.
A frozen BLIP captioner provides a label-free caption for each image, and the caption-guided enhancer uses cross-attention to align these instance-level cues with both support prototypes and query features.
As a result, PMCE moves from purely class-level semantics to a mixed granularity where category names govern prior selection and captions adjust feature representations on a per-image basis.
This helps the model capture contextual variations inside a class and strengthens discrimination in challenging 1-shot and coarse-grained settings.

Finally, PMCE keeps all large pre-trained components frozen.
The visual backbone is a standard ResNet-12 or Swin-T, and the only trainable parts are a lightweight enhancer and an auxiliary classifier on base classes.
Unlike recent works that fine-tune CLIP or pair FSL with very large VLM/LLM backbones, our improvements mainly stem from how semantics are used to select priors and reshape prototypes rather than from scaling model size.
This makes PMCE complementary to large-model pipelines and also easier to deploy in resource-constrained scenarios.

\cleardoublepage
\bibliographystyle{named}
\bibliography{ijcai26}

\begin{thebibliography}{}

\bibitem[\protect\citeauthoryear{Afrasiyabi \bgroup \em et al.\egroup }{2022}]{afrasiyabi2022matching}
Arman Afrasiyabi, Hugo Larochelle, Jean{-}Fran{\c{c}}ois Lalonde, and Christian Gagn{\'{e}}.
\newblock Matching feature sets for few-shot image classification.
\newblock In {\em {IEEE/CVF} Conference on Computer Vision and Pattern Recognition, {CVPR} 2022, New Orleans, LA, USA, June 18-24, 2022}, pages 9004--9014. {IEEE}, 2022.

\bibitem[\protect\citeauthoryear{Bertinetto \bgroup \em et al.\egroup }{2019}]{bertinetto2018meta}
Luca Bertinetto, Jo{\~{a}}o~F. Henriques, Philip H.~S. Torr, and Andrea Vedaldi.
\newblock Meta-learning with differentiable closed-form solvers.
\newblock In {\em 7th International Conference on Learning Representations, {ICLR} 2019, New Orleans, LA, USA, May 6-9, 2019}. OpenReview.net, 2019.

\bibitem[\protect\citeauthoryear{Cai \bgroup \em et al.\egroup }{2024a}]{DBLP:conf/ijcai/CaiLH024}
Hecheng Cai, Yang Liu, Shudong Huang, and Jiancheng Lv.
\newblock With a little help from language: Semantic enhanced visual prototype framework for few-shot learning.
\newblock In {\em Proceedings of the Thirty-Third International Joint Conference on Artificial Intelligence, {IJCAI} 2024, Jeju, South Korea, August 3-9, 2024}, pages 3751--3759. ijcai.org, 2024.

\bibitem[\protect\citeauthoryear{Cai \bgroup \em et al.\egroup }{2024b}]{cai2024little}
Hecheng Cai, Yang Liu, Shudong Huang, and Jiancheng Lv.
\newblock With a little help from language: Semantic enhanced visual prototype framework for few-shot learning.
\newblock In {\em Proceedings of the Thirty-Third International Joint Conference on Artificial Intelligence, {IJCAI} 2024, Jeju, South Korea, August 3-9, 2024}, pages 3751--3759. ijcai.org, 2024.

\bibitem[\protect\citeauthoryear{Changpinyo \bgroup \em et al.\egroup }{2021}]{changpinyo2021conceptual}
Soravit Changpinyo, Piyush Sharma, Nan Ding, and Radu Soricut.
\newblock Conceptual 12m: Pushing web-scale image-text pre-training to recognize long-tail visual concepts.
\newblock In {\em {IEEE} Conference on Computer Vision and Pattern Recognition, {CVPR} 2021, virtual, June 19-25, 2021}, pages 3558--3568. Computer Vision Foundation / {IEEE}, 2021.

\bibitem[\protect\citeauthoryear{Chen \bgroup \em et al.\egroup }{2019}]{chen2019closer}
Wei{-}Yu Chen, Yen{-}Cheng Liu, Zsolt Kira, Yu{-}Chiang~Frank Wang, and Jia{-}Bin Huang.
\newblock A closer look at few-shot classification.
\newblock In {\em 7th International Conference on Learning Representations, {ICLR} 2019, New Orleans, LA, USA, May 6-9, 2019}. OpenReview.net, 2019.

\bibitem[\protect\citeauthoryear{Chen \bgroup \em et al.\egroup }{2021a}]{chen2021metabaseline}
Yinbo Chen, Zhuang Liu, Huijuan Xu, Trevor Darrell, and Xiaolong Wang.
\newblock Meta-baseline: Exploring simple meta-learning for few-shot learning.
\newblock In {\em 2021 {IEEE/CVF} International Conference on Computer Vision, {ICCV} 2021, Montreal, QC, Canada, October 10-17, 2021}, pages 9042--9051. {IEEE}, 2021.

\bibitem[\protect\citeauthoryear{Chen \bgroup \em et al.\egroup }{2021b}]{chen2021semantics}
Zhi Chen, Yadan Luo, Ruihong Qiu, Sen Wang, Zi~Huang, Jingjing Li, and Zheng Zhang.
\newblock Semantics disentangling for generalized zero-shot learning.
\newblock In {\em 2021 {IEEE/CVF} International Conference on Computer Vision, {ICCV} 2021, Montreal, QC, Canada, October 10-17, 2021}, pages 8692--8700. {IEEE}, 2021.

\bibitem[\protect\citeauthoryear{Chen \bgroup \em et al.\egroup }{2023a}]{chen2023sparse}
Haoxing Chen, Huaxiong Li, Yaohui Li, and Chunlin Chen.
\newblock Sparse spatial transformers for few-shot learning.
\newblock {\em Sci. China Inf. Sci.}, 66(11), 2023.

\bibitem[\protect\citeauthoryear{Chen \bgroup \em et al.\egroup }{2023b}]{DBLP:journals/corr/abs-2303-14123}
Wentao Chen, Chenyang Si, Zhang Zhang, Liang Wang, Zilei Wang, and Tieniu Tan.
\newblock Semantic prompt for few-shot image recognition.
\newblock {\em CoRR}, abs/2303.14123, 2023.

\bibitem[\protect\citeauthoryear{Chen \bgroup \em et al.\egroup }{2023c}]{chen2022gsmflow}
Zhi Chen, Yadan Luo, Sen Wang, Jingjing Li, and Zi~Huang.
\newblock Gsmflow: Generation shifts mitigating flow for generalized zero-shot learning.
\newblock {\em {IEEE} Trans. Multim.}, 25:5374--5385, 2023.

\bibitem[\protect\citeauthoryear{Chen \bgroup \em et al.\egroup }{2023d}]{chen2023zero}
Zhi Chen, Peng{-}Fei Zhang, Jingjing Li, Sen Wang, and Zi~Huang.
\newblock Zero-shot learning by harnessing adversarial samples.
\newblock In Abdulmotaleb El{-}Saddik, Tao Mei, Rita Cucchiara, Marco Bertini, Diana Patricia~Tobon Vallejo, Pradeep~K. Atrey, and M.~Shamim Hossain, editors, {\em Proceedings of the 31st {ACM} International Conference on Multimedia, {MM} 2023, Ottawa, ON, Canada, 29 October 2023- 3 November 2023}, pages 4138--4146. {ACM}, 2023.

\bibitem[\protect\citeauthoryear{Chen \bgroup \em et al.\egroup }{2024}]{chen2024mutual}
Derong Chen, Feiyu Chen, Deqiang Ouyang, and Jie Shao.
\newblock Mutual correlation network for few-shot learning.
\newblock {\em Neural Networks}, 175:106289, 2024.

\bibitem[\protect\citeauthoryear{Chen \bgroup \em et al.\egroup }{2025a}]{chen2025ddc}
Lingxing Chen, Yang Gu, Yi~Guo, Fan Dong, Dongmei Jiang, and Yiqiang Chen.
\newblock {DDC:} dynamic distribution calibration for few-shot learning under multi-scale representation.
\newblock {\em Knowl. Based Syst.}, 311:113030, 2025.

\bibitem[\protect\citeauthoryear{Chen \bgroup \em et al.\egroup }{2025b}]{chen2025exploring}
Xingye Chen, Wenxiao Wu, Li~Ma, Xinge You, Changxin Gao, Nong Sang, and Yuanjie Shao.
\newblock Exploring sample relationship for few-shot classification.
\newblock {\em Pattern Recognit.}, 159:111089, 2025.

\bibitem[\protect\citeauthoryear{Chen \bgroup \em et al.\egroup }{2025c}]{chen2025svip}
Zhi Chen, Zecheng Zhao, Jingcai Guo, Jingjing Li, and Huang Zi.
\newblock Svip: Semantically contextualized visual patches for zero-shot learning.
\newblock In {\em Proceedings of the IEEE/CVF international conference on computer vision}, pages 3346--3356, 2025.

\bibitem[\protect\citeauthoryear{Chen \bgroup \em et al.\egroup }{2026}]{DBLP:journals/pr/ChenYTLH26}
Zhi Chen, Xin Yu, Xiaohui Tao, Yan Li, and Zi~Huang.
\newblock Cluster-aware prompt ensemble learning for few-shot vision-language model adaptation.
\newblock {\em Pattern Recognit.}, 172:112596, 2026.

\bibitem[\protect\citeauthoryear{Cheng \bgroup \em et al.\egroup }{2023}]{cheng2023frequency}
Hao Cheng, Siyuan Yang, Joey~Tianyi Zhou, Lanqing Guo, and Bihan Wen.
\newblock Frequency guidance matters in few-shot learning.
\newblock In {\em {IEEE/CVF} International Conference on Computer Vision, {ICCV} 2023, Paris, France, October 1-6, 2023}, pages 11780--11790. {IEEE}, 2023.

\bibitem[\protect\citeauthoryear{Cumplido \bgroup \em et al.\egroup }{2025}]{DBLP:conf/cvpr/Cumplido0R25}
Javier~R{\'{o}}denas Cumplido, Eduardo Aguilar, and Petia Radeva.
\newblock Slot attention-based feature filtering for few-shot learning.
\newblock In {\em {IEEE/CVF} Conference on Computer Vision and Pattern Recognition Workshops, {CVPR} Workshops 2025, Nashville, TN, USA, June 11-15, 2025}, pages 30--40. Computer Vision Foundation / {IEEE}, 2025.

\bibitem[\protect\citeauthoryear{Dong \bgroup \em et al.\egroup }{2025}]{dong2025prsn}
Mengping Dong, Fei Li, Zhenbo Li, and Xue Liu.
\newblock {PRSN:} prototype resynthesis network with cross-image semantic alignment for few-shot image classification.
\newblock {\em Pattern Recognit.}, 159:111122, 2025.

\bibitem[\protect\citeauthoryear{Finn \bgroup \em et al.\egroup }{2017}]{finn2017model}
Chelsea Finn, Pieter Abbeel, and Sergey Levine.
\newblock Model-agnostic meta-learning for fast adaptation of deep networks.
\newblock In Doina Precup and Yee~Whye Teh, editors, {\em Proceedings of the 34th International Conference on Machine Learning, {ICML} 2017, Sydney, NSW, Australia, 6-11 August 2017}, volume~70 of {\em Proceedings of Machine Learning Research}, pages 1126--1135. {PMLR}, 2017.

\bibitem[\protect\citeauthoryear{Fu \bgroup \em et al.\egroup }{2023}]{fu2023styleadv}
Yuqian Fu, Yu~Xie, Yanwei Fu, and Yu{-}Gang Jiang.
\newblock Styleadv: Meta style adversarial training for cross-domain few-shot learning.
\newblock In {\em {IEEE/CVF} Conference on Computer Vision and Pattern Recognition, {CVPR} 2023, Vancouver, BC, Canada, June 17-24, 2023}, pages 24575--24584. {IEEE}, 2023.

\bibitem[\protect\citeauthoryear{Fu \bgroup \em et al.\egroup }{2025}]{fu2024prototype}
Meijun Fu, Xiaomin Wang, Jun Wang, and Zhang Yi.
\newblock Prototype bayesian meta-learning for few-shot image classification.
\newblock {\em {IEEE} Trans. Neural Networks Learn. Syst.}, 36(4):7010--7024, 2025.

\bibitem[\protect\citeauthoryear{Guo and Guo}{2020}]{guo2020novel}
Jingcai Guo and Song Guo.
\newblock A novel perspective to zero-shot learning: Towards an alignment of manifold structures via semantic feature expansion.
\newblock {\em IEEE Transactions on Multimedia}, 23:524--537, 2020.

\bibitem[\protect\citeauthoryear{Guo \bgroup \em et al.\egroup }{2021}]{guo2021conservative}
Jingcai Guo, Song Guo, Shiheng Ma, Yuxia Sun, and Yuanyuan Xu.
\newblock Conservative novelty synthesizing network for malware recognition in an open-set scenario.
\newblock {\em IEEE Transactions on Neural Networks and Learning Systems}, 34(2):662--676, 2021.

\bibitem[\protect\citeauthoryear{Guo \bgroup \em et al.\egroup }{2023}]{guo2023graph}
Jingcai Guo, Song Guo, Qihua Zhou, Ziming Liu, Xiaocheng Lu, and Fushuo Huo.
\newblock Graph knows unknowns: Reformulate zero-shot learning as sample-level graph recognition.
\newblock In {\em Proceedings of the AAAI conference on artificial intelligence}, pages 7775--7783. {AAAI} Press, 2023.

\bibitem[\protect\citeauthoryear{Guo \bgroup \em et al.\egroup }{2024a}]{guo2024fine}
Jingcai Guo, Zhijie Rao, Zhi Chen, Jingren Zhou, and Dacheng Tao.
\newblock Fine-grained zero-shot learning: Advances, challenges, and prospects.
\newblock {\em arXiv preprint arXiv:2401.17766}, 2024.

\bibitem[\protect\citeauthoryear{Guo \bgroup \em et al.\egroup }{2024b}]{guo2024multimodal}
Jingcai Guo, Han Wang, Yuanyuan Xu, Wenchao Xu, Yufeng Zhan, Yuxia Sun, and Song Guo.
\newblock Multimodal dual-embedding networks for malware open-set recognition.
\newblock {\em IEEE Transactions on Neural Networks and Learning Systems}, 36(3):4545--4559, 2024.

\bibitem[\protect\citeauthoryear{Guo \bgroup \em et al.\egroup }{2024c}]{guo2024parsnets}
Jingcai Guo, Qihua Zhou, Xiaocheng Lu, Ruibin Li, Ziming Liu, Jie Zhang, Bo~Han, Junyang Chen, Xin Xie, and Song Guo.
\newblock Parsnets: A parsimonious composition of orthogonal and low-rank linear networks for zero-shot learning.
\newblock In {\em Proceedings of the Thirty-Third International Joint Conference on Artificial Intelligence, IJCAI-24}, pages 4062--4070, 2024.

\bibitem[\protect\citeauthoryear{Hao \bgroup \em et al.\egroup }{2023}]{hao2023cpea}
Fusheng Hao, Fengxiang He, Liu Liu, Fuxiang Wu, Dacheng Tao, and Jun Cheng.
\newblock Class-aware patch embedding adaptation for few-shot image classification.
\newblock In {\em {IEEE/CVF} International Conference on Computer Vision, {ICCV} 2023, Paris, France, October 1-6, 2023}, pages 18859--18869. {IEEE}, 2023.

\bibitem[\protect\citeauthoryear{He \bgroup \em et al.\egroup }{2016}]{he2016deep}
Kaiming He, Xiangyu Zhang, Shaoqing Ren, and Jian Sun.
\newblock Deep residual learning for image recognition.
\newblock In {\em Proceedings of the IEEE conference on computer vision and pattern recognition}, pages 770--778, 2016.

\bibitem[\protect\citeauthoryear{He \bgroup \em et al.\egroup }{2023}]{He_2023_WACV}
Ju~He, Adam Kortylewski, and Alan Yuille.
\newblock Corl: Compositional representation learning for few-shot classification.
\newblock In {\em Proceedings of the IEEE/CVF Winter Conference on Applications of Computer Vision (WACV)}, pages 3890--3899, January 2023.

\bibitem[\protect\citeauthoryear{Hu and Ma}{2022}]{hu2022adversarial}
Yanxu Hu and Andy~J. Ma.
\newblock Adversarial feature augmentation for cross-domain few-shot classification.
\newblock In Shai Avidan, Gabriel~J. Brostow, Moustapha Ciss{\'{e}}, Giovanni~Maria Farinella, and Tal Hassner, editors, {\em Computer Vision - {ECCV} 2022 - 17th European Conference, Tel Aviv, Israel, October 23-27, 2022, Proceedings, Part {XX}}, volume 13680 of {\em Lecture Notes in Computer Science}, pages 20--37. Springer, 2022.

\bibitem[\protect\citeauthoryear{Krishna \bgroup \em et al.\egroup }{2017}]{krishna2017visual}
Ranjay Krishna, Yuke Zhu, Oliver Groth, Justin Johnson, Kenji Hata, Joshua Kravitz, Stephanie Chen, Yannis Kalantidis, Li{-}Jia Li, David~A. Shamma, Michael~S. Bernstein, and Li~Fei{-}Fei.
\newblock Visual genome: Connecting language and vision using crowdsourced dense image annotations.
\newblock {\em Int. J. Comput. Vis.}, 123(1):32--73, 2017.

\bibitem[\protect\citeauthoryear{Lee \bgroup \em et al.\egroup }{2019}]{lee2019meta}
Kwonjoon Lee, Subhransu Maji, Avinash Ravichandran, and Stefano Soatto.
\newblock Meta-learning with differentiable convex optimization.
\newblock In {\em {IEEE} Conference on Computer Vision and Pattern Recognition, {CVPR} 2019, Long Beach, CA, USA, June 16-20, 2019}, pages 10657--10665. Computer Vision Foundation / {IEEE}, 2019.

\bibitem[\protect\citeauthoryear{Li and Wang}{2023}]{DBLP:conf/mm/LiW23}
Zhuoling Li and Yong Wang.
\newblock Better integrating vision and semantics for improving few-shot classification.
\newblock In Abdulmotaleb El{-}Saddik, Tao Mei, Rita Cucchiara, Marco Bertini, Diana Patricia~Tobon Vallejo, Pradeep~K. Atrey, and M.~Shamim Hossain, editors, {\em Proceedings of the 31st {ACM} International Conference on Multimedia, {MM} 2023, Ottawa, ON, Canada, 29 October 2023- 3 November 2023}, pages 4737--4746. {ACM}, 2023.

\bibitem[\protect\citeauthoryear{Li \bgroup \em et al.\egroup }{2022}]{li2022blip}
Junnan Li, Dongxu Li, Caiming Xiong, and Steven C.~H. Hoi.
\newblock {BLIP:} bootstrapping language-image pre-training for unified vision-language understanding and generation.
\newblock In Kamalika Chaudhuri, Stefanie Jegelka, Le~Song, Csaba Szepesv{\'{a}}ri, Gang Niu, and Sivan Sabato, editors, {\em International Conference on Machine Learning, {ICML} 2022, 17-23 July 2022, Baltimore, Maryland, {USA}}, volume 162 of {\em Proceedings of Machine Learning Research}, pages 12888--12900. {PMLR}, 2022.

\bibitem[\protect\citeauthoryear{Li \bgroup \em et al.\egroup }{2025}]{DBLP:journals/corr/abs-2509-25033}
Wenhao Li, Qiangchang Wang, Xianjing Meng, Zhibin Wu, and Yilong Yin.
\newblock {VT-FSL:} bridging vision and text with llms for few-shot learning.
\newblock {\em CoRR}, abs/2509.25033, 2025.

\bibitem[\protect\citeauthoryear{Lin \bgroup \em et al.\egroup }{2014}]{lin2014microsoft}
Tsung{-}Yi Lin, Michael Maire, Serge~J. Belongie, James Hays, Pietro Perona, Deva Ramanan, Piotr Doll{\'{a}}r, and C.~Lawrence Zitnick.
\newblock Microsoft {COCO:} common objects in context.
\newblock In David~J. Fleet, Tom{\'{a}}s Pajdla, Bernt Schiele, and Tinne Tuytelaars, editors, {\em Computer Vision - {ECCV} 2014 - 13th European Conference, Zurich, Switzerland, September 6-12, 2014, Proceedings, Part {V}}, volume 8693 of {\em Lecture Notes in Computer Science}, pages 740--755. Springer, 2014.

\bibitem[\protect\citeauthoryear{Liu \bgroup \em et al.\egroup }{2021}]{liu2021swin}
Ze~Liu, Yutong Lin, Yue Cao, Han Hu, Yixuan Wei, Zheng Zhang, Stephen Lin, and Baining Guo.
\newblock Swin transformer: Hierarchical vision transformer using shifted windows.
\newblock In {\em 2021 {IEEE/CVF} International Conference on Computer Vision, {ICCV} 2021, Montreal, QC, Canada, October 10-17, 2021}, pages 9992--10002. {IEEE}, 2021.

\bibitem[\protect\citeauthoryear{Liu \bgroup \em et al.\egroup }{2023}]{liu2023capturing}
Xinyue Liu, Ligang Liu, Han Liu, and Xiaotong Zhang.
\newblock Capturing the few-shot class distribution: Transductive distribution optimization.
\newblock {\em Pattern Recognit.}, 138:109371, 2023.

\bibitem[\protect\citeauthoryear{Lu \bgroup \em et al.\egroup }{2023}]{lu2023decomposed}
Xiaocheng Lu, Song Guo, Ziming Liu, and Jingcai Guo.
\newblock Decomposed soft prompt guided fusion enhancing for compositional zero-shot learning.
\newblock In {\em Proceedings of the IEEE/CVF Conference on Computer Vision and Pattern Recognition}, pages 23560--23569, 2023.

\bibitem[\protect\citeauthoryear{Luo \bgroup \em et al.\egroup }{2023}]{luo2023closeragain}
Xu~Luo, Hao Wu, Ji~Zhang, Lianli Gao, Jing Xu, and Jingkuan Song.
\newblock A closer look at few-shot classification again.
\newblock In Andreas Krause, Emma Brunskill, Kyunghyun Cho, Barbara Engelhardt, Sivan Sabato, and Jonathan Scarlett, editors, {\em International Conference on Machine Learning, {ICML} 2023, 23-29 July 2023, Honolulu, Hawaii, {USA}}, volume 202 of {\em Proceedings of Machine Learning Research}, pages 23103--23123. {PMLR}, 2023.

\bibitem[\protect\citeauthoryear{Ordonez \bgroup \em et al.\egroup }{2011}]{ordonez2011im2text}
Vicente Ordonez, Girish Kulkarni, and Tamara~L. Berg.
\newblock Im2text: Describing images using 1 million captioned photographs.
\newblock In John Shawe{-}Taylor, Richard~S. Zemel, Peter~L. Bartlett, Fernando C.~N. Pereira, and Kilian~Q. Weinberger, editors, {\em Advances in Neural Information Processing Systems 24: 25th Annual Conference on Neural Information Processing Systems 2011. Proceedings of a meeting held 12-14 December 2011, Granada, Spain}, pages 1143--1151, 2011.

\bibitem[\protect\citeauthoryear{Oreshkin \bgroup \em et al.\egroup }{2018}]{oreshkin2018tadam}
Boris~N. Oreshkin, Pau~Rodr{\'{\i}}guez L{\'{o}}pez, and Alexandre Lacoste.
\newblock {TADAM:} task dependent adaptive metric for improved few-shot learning.
\newblock In Samy Bengio, Hanna~M. Wallach, Hugo Larochelle, Kristen Grauman, Nicol{\`{o}} Cesa{-}Bianchi, and Roman Garnett, editors, {\em Advances in Neural Information Processing Systems 31: Annual Conference on Neural Information Processing Systems 2018, NeurIPS 2018, December 3-8, 2018, Montr{\'{e}}al, Canada}, pages 719--729, 2018.

\bibitem[\protect\citeauthoryear{Peng \bgroup \em et al.\egroup }{2019}]{peng2019few}
Zhimao Peng, Zechao Li, Junge Zhang, Yan Li, Guo{-}Jun Qi, and Jinhui Tang.
\newblock Few-shot image recognition with knowledge transfer.
\newblock In {\em 2019 {IEEE/CVF} International Conference on Computer Vision, {ICCV} 2019, Seoul, Korea (South), October 27 - November 2, 2019}, pages 441--449. {IEEE}, 2019.

\bibitem[\protect\citeauthoryear{Radford \bgroup \em et al.\egroup }{2021}]{radford2021learning}
Alec Radford, Jong~Wook Kim, Chris Hallacy, Aditya Ramesh, Gabriel Goh, Sandhini Agarwal, Girish Sastry, Amanda Askell, Pamela Mishkin, Jack Clark, Gretchen Krueger, and Ilya Sutskever.
\newblock Learning transferable visual models from natural language supervision.
\newblock In Marina Meila and Tong Zhang, editors, {\em Proceedings of the 38th International Conference on Machine Learning, {ICML} 2021, 18-24 July 2021, Virtual Event}, volume 139 of {\em Proceedings of Machine Learning Research}, pages 8748--8763. {PMLR}, 2021.

\bibitem[\protect\citeauthoryear{Ren \bgroup \em et al.\egroup }{2018}]{ren2018meta}
Mengye Ren, Eleni Triantafillou, Sachin Ravi, Jake Snell, Kevin Swersky, Joshua~B. Tenenbaum, Hugo Larochelle, and Richard~S. Zemel.
\newblock Meta-learning for semi-supervised few-shot classification.
\newblock In {\em 6th International Conference on Learning Representations, {ICLR} 2018, Vancouver, BC, Canada, April 30 - May 3, 2018, Conference Track Proceedings}. OpenReview.net, 2018.

\bibitem[\protect\citeauthoryear{Russakovsky \bgroup \em et al.\egroup }{2015}]{russakovsky2015imagenet}
Olga Russakovsky, Jia Deng, Hao Su, Jonathan Krause, Sanjeev Satheesh, Sean Ma, Zhiheng Huang, Andrej Karpathy, Aditya Khosla, Michael~S. Bernstein, Alexander~C. Berg, and Li~Fei{-}Fei.
\newblock Imagenet large scale visual recognition challenge.
\newblock {\em Int. J. Comput. Vis.}, 115(3):211--252, 2015.

\bibitem[\protect\citeauthoryear{Schuhmann \bgroup \em et al.\egroup }{2021}]{DBLP:journals/corr/abs-2111-02114}
Christoph Schuhmann, Richard Vencu, Romain Beaumont, Robert Kaczmarczyk, Clayton Mullis, Aarush Katta, Theo Coombes, Jenia Jitsev, and Aran Komatsuzaki.
\newblock {LAION-400M:} open dataset of clip-filtered 400 million image-text pairs.
\newblock {\em CoRR}, abs/2111.02114, 2021.

\bibitem[\protect\citeauthoryear{Snell \bgroup \em et al.\egroup }{2017}]{snell2017prototypical}
Jake Snell, Kevin Swersky, and Richard~S. Zemel.
\newblock Prototypical networks for few-shot learning.
\newblock In Isabelle Guyon, Ulrike von Luxburg, Samy Bengio, Hanna~M. Wallach, Rob Fergus, S.~V.~N. Vishwanathan, and Roman Garnett, editors, {\em Advances in Neural Information Processing Systems 30: Annual Conference on Neural Information Processing Systems 2017, December 4-9, 2017, Long Beach, CA, {USA}}, pages 4077--4087, 2017.

\bibitem[\protect\citeauthoryear{Sung \bgroup \em et al.\egroup }{2018}]{sung2018learning}
Flood Sung, Yongxin Yang, Li~Zhang, Tao Xiang, Philip H.~S. Torr, and Timothy~M. Hospedales.
\newblock Learning to compare: Relation network for few-shot learning.
\newblock In {\em 2018 {IEEE} Conference on Computer Vision and Pattern Recognition, {CVPR} 2018, Salt Lake City, UT, USA, June 18-22, 2018}, pages 1199--1208. Computer Vision Foundation / {IEEE} Computer Society, 2018.

\bibitem[\protect\citeauthoryear{Tang \bgroup \em et al.\egroup }{2025}]{DBLP:conf/ijcai/TangHQ25}
Hao Tang, Shengfeng He, and Jing Qin.
\newblock Connecting giants: Synergistic knowledge transfer of large multimodal models for few-shot learning.
\newblock In {\em Proceedings of the Thirty-Fourth International Joint Conference on Artificial Intelligence, {IJCAI} 2025, Montreal, Canada, August 16-22, 2025}, pages 6227--6235. ijcai.org, 2025.

\bibitem[\protect\citeauthoryear{Tian \bgroup \em et al.\egroup }{2020}]{tian2020rethinking}
Yonglong Tian, Yue Wang, Dilip Krishnan, Joshua~B. Tenenbaum, and Phillip Isola.
\newblock Rethinking few-shot image classification: {A} good embedding is all you need?
\newblock In Andrea Vedaldi, Horst Bischof, Thomas Brox, and Jan{-}Michael Frahm, editors, {\em Computer Vision - {ECCV} 2020 - 16th European Conference, Glasgow, UK, August 23-28, 2020, Proceedings, Part {XIV}}, volume 12359 of {\em Lecture Notes in Computer Science}, pages 266--282. Springer, 2020.

\bibitem[\protect\citeauthoryear{Tseng \bgroup \em et al.\egroup }{2020}]{DBLP:conf/iclr/TsengLH020}
Hung{-}Yu Tseng, Hsin{-}Ying Lee, Jia{-}Bin Huang, and Ming{-}Hsuan Yang.
\newblock Cross-domain few-shot classification via learned feature-wise transformation.
\newblock In {\em 8th International Conference on Learning Representations, {ICLR} 2020, Addis Ababa, Ethiopia, April 26-30, 2020}. OpenReview.net, 2020.

\bibitem[\protect\citeauthoryear{Vinyals \bgroup \em et al.\egroup }{2016}]{vinyals2016matching}
Oriol Vinyals, Charles Blundell, Tim Lillicrap, Koray Kavukcuoglu, and Daan Wierstra.
\newblock Matching networks for one shot learning.
\newblock In Daniel~D. Lee, Masashi Sugiyama, Ulrike von Luxburg, Isabelle Guyon, and Roman Garnett, editors, {\em Advances in Neural Information Processing Systems 29: Annual Conference on Neural Information Processing Systems 2016, December 5-10, 2016, Barcelona, Spain}, pages 3630--3638, 2016.

\bibitem[\protect\citeauthoryear{Wang \bgroup \em et al.\egroup }{2019}]{wang2019simpleshot}
Yan Wang, Wei{-}Lun Chao, Kilian~Q. Weinberger, and Laurens van~der Maaten.
\newblock Simpleshot: Revisiting nearest-neighbor classification for few-shot learning.
\newblock {\em CoRR}, abs/1911.04623, 2019.

\bibitem[\protect\citeauthoryear{Wang \bgroup \em et al.\egroup }{2023a}]{wang2023visual}
Runqi Wang, Hao Zheng, Xiaoyue Duan, Jianzhuang Liu, Yuning Lu, Tian Wang, Songcen Xu, and Baochang Zhang.
\newblock Few-shot learning with visual distribution calibration and cross-modal distribution alignment.
\newblock In {\em {IEEE/CVF} Conference on Computer Vision and Pattern Recognition, {CVPR} 2023, Vancouver, BC, Canada, June 17-24, 2023}, pages 23445--23454. {IEEE}, 2023.

\bibitem[\protect\citeauthoryear{Wang \bgroup \em et al.\egroup }{2023b}]{wang2023few}
Xixi Wang, Xiao Wang, Bo~Jiang, and Bin Luo.
\newblock Few-shot learning meets transformer: Unified query-support transformers for few-shot classification.
\newblock {\em {IEEE} Trans. Circuits Syst. Video Technol.}, 33(12):7789--7802, 2023.

\bibitem[\protect\citeauthoryear{Wu and Hu}{2022}]{wu2022improved}
Jiaying Wu and Jinglu Hu.
\newblock Improved prior selection using semantics in maximum a posteriori for few-shot learning.
\newblock {\em Knowl. Based Syst.}, 237:107688, 2022.

\bibitem[\protect\citeauthoryear{Wu \bgroup \em et al.\egroup }{2021}]{wu2021feature}
Jiaying Wu, Ning Dong, Fan Liu, Sai Yang, and Jinglu Hu.
\newblock Feature hallucination via maximum {A} posteriori for few-shot learning.
\newblock {\em Knowl. Based Syst.}, 225:107129, 2021.

\bibitem[\protect\citeauthoryear{Xing \bgroup \em et al.\egroup }{2019}]{xing2019adaptive}
Chen Xing, Negar Rostamzadeh, Boris~N. Oreshkin, and Pedro~O. Pinheiro.
\newblock Adaptive cross-modal few-shot learning.
\newblock In Hanna~M. Wallach, Hugo Larochelle, Alina Beygelzimer, Florence d'Alch{\'{e}}{-}Buc, Emily~B. Fox, and Roman Garnett, editors, {\em Advances in Neural Information Processing Systems 32: Annual Conference on Neural Information Processing Systems 2019, NeurIPS 2019, December 8-14, 2019, Vancouver, BC, Canada}, pages 4848--4858, 2019.

\bibitem[\protect\citeauthoryear{Yang \bgroup \em et al.\egroup }{2021}]{yang2021free}
Shuo Yang, Lu~Liu, and Min Xu.
\newblock Free lunch for few-shot learning: Distribution calibration.
\newblock In {\em 9th International Conference on Learning Representations, {ICLR} 2021, Virtual Event, Austria, May 3-7, 2021}. OpenReview.net, 2021.

\bibitem[\protect\citeauthoryear{Yang \bgroup \em et al.\egroup }{2022a}]{yang2022sega}
Fengyuan Yang, Ruiping Wang, and Xilin Chen.
\newblock {SEGA:} semantic guided attention on visual prototype for few-shot learning.
\newblock In {\em {IEEE/CVF} Winter Conference on Applications of Computer Vision, {WACV} 2022, Waikoloa, HI, USA, January 3-8, 2022}, pages 1586--1596. {IEEE}, 2022.

\bibitem[\protect\citeauthoryear{Yang \bgroup \em et al.\egroup }{2022b}]{yang2021distcal}
Shuo Yang, Songhua Wu, Tongliang Liu, and Min Xu.
\newblock Bridging the gap between few-shot and many-shot learning via distribution calibration.
\newblock {\em {IEEE} Trans. Pattern Anal. Mach. Intell.}, 44(12):9830--9843, 2022.

\bibitem[\protect\citeauthoryear{Yang \bgroup \em et al.\egroup }{2023}]{Yang_2023_WACV}
Fengyuan Yang, Ruiping Wang, and Xilin Chen.
\newblock Semantic guided latent parts embedding for few-shot learning.
\newblock In {\em {IEEE/CVF} Winter Conference on Applications of Computer Vision, {WACV} 2023, Waikoloa, HI, USA, January 2-7, 2023}, pages 5436--5446. {IEEE}, 2023.

\bibitem[\protect\citeauthoryear{Yue \bgroup \em et al.\egroup }{2025}]{yue2025multi}
Ling Yue, Lin Feng, Qiuping Shuai, Zihao Li, and Lingxiao Xu.
\newblock Multi-level adaptive feature representation based on task augmentation for cross-domain few-shot learning.
\newblock {\em Appl. Intell.}, 55(4):291, 2025.

\bibitem[\protect\citeauthoryear{Zhang \bgroup \em et al.\egroup }{2024a}]{zhang2024semfew}
Hai Zhang, Junzhe Xu, Shanlin Jiang, and Zhenan He.
\newblock Simple semantic-aided few-shot learning.
\newblock In {\em {IEEE/CVF} Conference on Computer Vision and Pattern Recognition, {CVPR} 2024, Seattle, WA, USA, June 16-22, 2024}, pages 28588--28597. {IEEE}, 2024.

\bibitem[\protect\citeauthoryear{Zhang \bgroup \em et al.\egroup }{2024b}]{Zhang2024Exploring}
Tiange Zhang, Qing Cai, Feng Gao, Lin Qi, and Junyu Dong.
\newblock Exploring cross-domain few-shot classification via frequency-aware prompting.
\newblock In {\em Proceedings of the Thirty-Third International Joint Conference on Artificial Intelligence, {IJCAI} 2024, Jeju, South Korea, August 3-9, 2024}, pages 5490--5498. ijcai.org, 2024.

\bibitem[\protect\citeauthoryear{Zhang \bgroup \em et al.\egroup }{2025}]{zhang2024few}
Yourun Zhang, Maoguo Gong, Jianzhao Li, Kaiyuan Feng, and Mingyang Zhang.
\newblock Few-shot learning with enhancements to data augmentation and feature extraction.
\newblock {\em {IEEE} Trans. Neural Networks Learn. Syst.}, 36(4):6655--6668, 2025.

\bibitem[\protect\citeauthoryear{Zhao \bgroup \em et al.\egroup }{2024}]{zhao2024multi}
Peng Zhao, Yin Wang, Wei Wang, Jie Mu, Huiting Liu, Cong Wang, and Xiaochun Cao.
\newblock Multi-attention based visual-semantic interaction for few-shot learning.
\newblock In {\em Proceedings of the Thirty-Third International Joint Conference on Artificial Intelligence, {IJCAI} 2024, Jeju, South Korea, August 3-9, 2024}, pages 1753--1761. ijcai.org, 2024.

\bibitem[\protect\citeauthoryear{Zhou \bgroup \em et al.\egroup }{2023}]{zhou2023revisiting}
Fei Zhou, Peng Wang, Lei Zhang, Wei Wei, and Yanning Zhang.
\newblock Revisiting prototypical network for cross domain few-shot learning.
\newblock In {\em {IEEE/CVF} Conference on Computer Vision and Pattern Recognition, {CVPR} 2023, Vancouver, BC, Canada, June 17-24, 2023}, pages 20061--20070. {IEEE}, 2023.

\bibitem[\protect\citeauthoryear{Zhou \bgroup \em et al.\egroup }{2025}]{zhou2025less}
Chunpeng Zhou, Zhi Yu, Xilu Yuan, Sheng Zhou, Jiajun Bu, and Haishuai Wang.
\newblock Less is more: {A} closer look at semantic-based few-shot learning.
\newblock {\em Inf. Fusion}, 114:102672, 2025.

\bibitem[\protect\citeauthoryear{Zou \bgroup \em et al.\egroup }{2024}]{zou2024flatten}
Yixiong Zou, Yicong Liu, Yiman Hu, Yuhua Li, and Ruixuan Li.
\newblock Flatten long-range loss landscapes for cross-domain few-shot learning.
\newblock In {\em {IEEE/CVF} Conference on Computer Vision and Pattern Recognition, {CVPR} 2024, Seattle, WA, USA, June 16-22, 2024}, pages 23575--23584. {IEEE}, 2024.

\end{thebibliography}

% \cleardoublepage
% \appendix 
\end{document}